%% file: main.tex
\documentclass[sn-aps]{sn-jnl}

\usepackage{booktabs}
\usepackage{longtable}
\usepackage{graphicx}%
\usepackage{multirow}%
\usepackage{amsmath,amssymb,amsfonts}%
\usepackage{amsthm}%
\usepackage{mathrsfs}%
\usepackage[title]{appendix}%
\usepackage{xcolor}%
\usepackage{textcomp}%
\usepackage{manyfoot}%
\usepackage{booktabs}%
\usepackage{algorithm}%
\usepackage{listings}%
\usepackage{float}
\usepackage{amssymb}

\usepackage{rotating}
\usepackage{pdflscape}
\usepackage{multirow} 
\usepackage{array}    
 \usepackage{comment}
\usepackage{booktabs} 
\usepackage{pifont}
\newcommand{\xmark}{\ding{55}} 

\theoremstyle{thmstyleone}%
\theoremstyle{thmstyletwo}%

\theoremstyle{thmstylethree}%

\begin{document}

\title{Data Authorisation and Validation in Autonomous Vehicles: A Critical Review \thanks{This is the author version}} 


\author*{\fnm{Reem} \sur{Alhabib}}\email{reem.alhabib@york.ac.uk}

\author*{\fnm{Poonam} \sur{Yadav}}\email{poonam.yadav@york.ac.uk}

\affil{\orgdiv{Computer Science}, \orgname{University of York}}


\abstract{Autonomous Vehicles (AVs) are becoming increasingly prevalent due to their potential to improve road safety and reduce environmental impact. These vehicles rely on Automated Driving Systems (ADS), which integrate multiple sensors and actuators. While some AVs operate with minimal human intervention, fully autonomous systems eliminate the need for human control entirely. Despite advances in AV technologies, secure and trustworthy data management remains a significant challenge.

This survey focuses on two relatively underexplored aspects in AV environments: data authorisation and validation. It examines the key related challenges and reviews existing solutions. The findings highlight critical gaps in current approaches and suggest future research directions to enhance AV data authorisation and validation. }

\keywords{Autonomous Vehicle, Automated Driving System, Data Authorisation, Data Validation}



\maketitle

\noindent\textit{This is the author version}

\newcommand{\dashitem}{%
 \renewcommand{\labelitemi}{--}%
  \item
  \renewcommand{\labelitemi}{$\bullet$}%
  }

\input{Section1_Introduction}

\input{Section2_Background}

\input{Section3_dataflow}

\input{Section4_concerens}

\input{Section5_results_and_conclusion}

\bibliography{sn-bibliography}

\end{document}

%% file: Section1_Introduction.tex
\section{Introduction} \label{Introduction}

AVs are an innovation in the automotive sector as they provide safer, more effective, and ecologically friendly transportation options \cite{papadoulis2019evaluating, pan2021shared}. In addition, they have the potential to significantly contribute to economic growth through various aspects. For example, they can substantially decrease the number of traffic accidents, leading to considerable economic savings \cite{winston2020autonomous}. Moreover, their impact on land use is notable, as repurposing parking areas for the real estate industry can increase land value by 5\% \cite{clements2017economic, othman2022exploring}. Accordingly, in recent years, prominent automotive manufacturers have invested substantially in a diverse range of AV technologies, amounting to billions of dollars. Several sources estimate that the AV market share will take between 15 and 20 years to reach 25 percent globally~\cite{liu2022business}. In line with this trend and industrial development, governments worldwide provide guidance that allows and encourages on-road AV trials. For example, the UK published a code of practice (2019) that specifies certain road trials that include various levels of automation~\cite{UKCode2019}. Subsequently, in early 2022, the government changed the Highway Code to ensure the first self-driving vehicles are introduced safely on the roads~\cite{gov}.

The Automated Driving System (ADS) is an integrated vehicle system that utilises various in-vehicle technologies and sensors to navigate autonomously from a starting point to a predefined destination. It comprises multiple control units that design a system to complete all driving tasks without human intervention. Cameras, GPS, and other sensors are connected to exchange data to facilitate independent driving decisions. 
These systems operate in a dynamic environment that demands real-time, rapid data feeding from various sources, including external roadsides and other vehicles, to the onboard sensors, which need to make continuous control decisions. Ensuring trust in automated driving systems relies heavily on the integrity and reliability of the surrounding data ecosystem. Thus, optimising data use is essential to improve the functionality of autonomous car systems. However, data collection, generation, processing, and storage challenges present critical research areas. Issues such as data privacy, integrity, and accessibility must be addressed to ensure reliable decision-making in real-time. As the complexity of these data interactions increases, innovative solutions are required to manage and safeguard the vast amounts of information generated by ADS.

While previous surveys have explored data security and privacy in AV systems, they often lack a dedicated focus on authorisation and validation mechanisms or discuss these aspects only at specific stages rather than across the entire data lifecycle. This paper critically examines the data aspects of AVs and provides an overview of ADS technology. In particular, it offers important insights into the definition, structure, data flow, ownership dynamics, and difficulties associated with AVs. Its systematic method strengthens credibility and advances knowledge of the consequences of AV data management.
\subsection{Contribution}
This paper aims to provide a comprehensive understanding of data and information flow in AV with a particular focus on authorisation and validation challenges across different data processing stages. Unlike previous surveys that primarily examine security, privacy, or general validation frameworks, our study provides a stage-specific analysis of data authorisation and validation challenges. As shown in Table \ref{survey_compirsion}, existing surveys often focus on specific aspects, such as security risks, blockchain-based access control, or legal considerations, while our work systematically categorises these challenges based on the data flow stages: collection, transmission, processing,  actuation and storage.

\subsection{Paper Structure}
In this paper, section~\ref{background} provides a brief overview of ADS, describing the information required to understand the structure and installation of these vehicles and all related technologies, including an overview of ADS's potential benefits and costs. It also introduces key data authorisation and validation concepts, explaining their roles in ensuring secure and accurate data handling. Section~\ref{Dataflow} discusses various data management issues, with each subsection addressing the requirements, current problems, and gaps and presenting solutions for each data lifecycle stage. Section~\ref{Concerns} reviews other related concerns, such as security, privacy, and ethics. Section~\ref{Future} considers open questions and future work.

\subsection{Methodology}
\begin{sidewaysfigure}
\centering
\includegraphics[height=0.4\textheight]{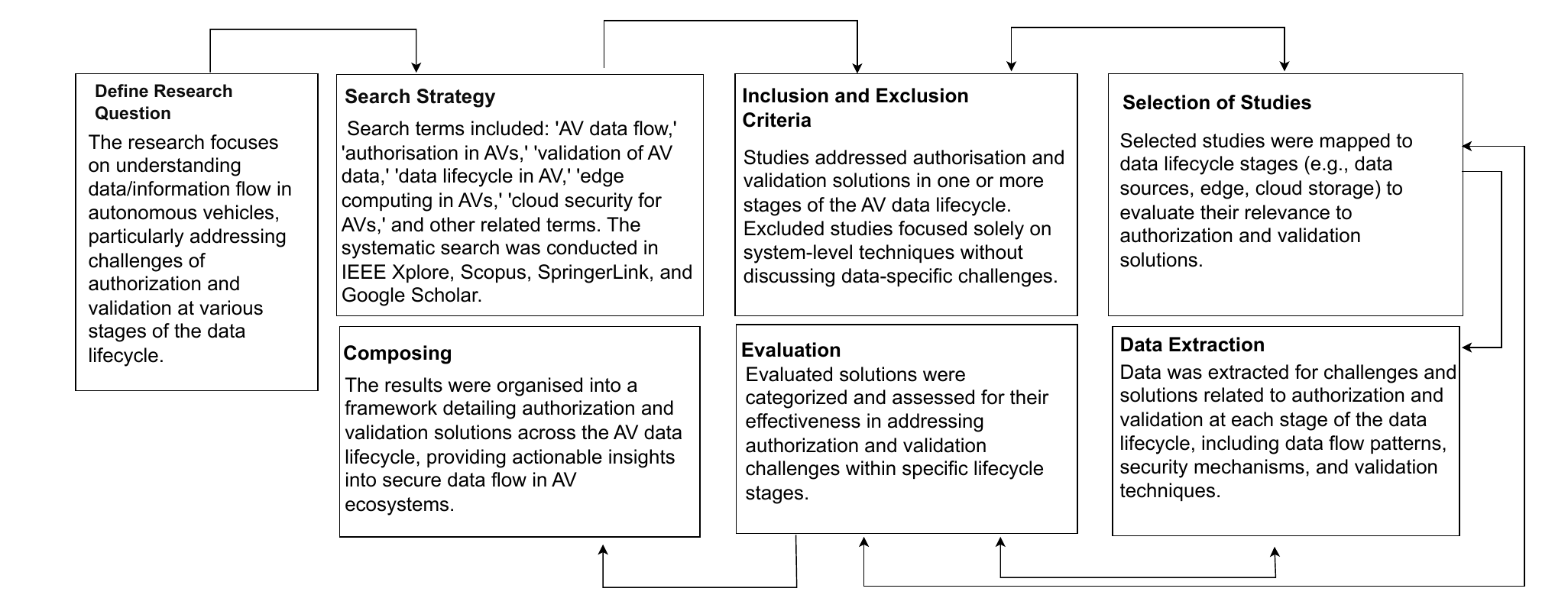}
\caption{\textbf{Review Methodology.} The flowchart outlines the key steps involved in conducting a systematic review of existing literature on AV. Double-headed arrows between some steps indicate bidirectional movement, highlighting the repetition of certain steps.}
\label{Method}
\end{sidewaysfigure}

This section describes, as in Figure \ref{Method}, the methodical review strategy used to make the search for and choosing a review strategy transparent and clear. The methodology of this survey follows a systematic approach aimed at understanding authorisation and validation within the data lifecycle of autonomous vehicles. It begins with a focus on identifying relevant studies through specific inclusion and exclusion criteria, ensuring that only those addressing data-specific challenges are considered. In addition, selected studies were mapped to various lifecycle stages (such as data sources, edge computing, and cloud storage) to evaluate their impact on authorisation and validation solutions.  A systematic search was conducted in major academic databases using targeted keywords to gather relevant literature.

\begin{longtable}{|p{1.0cm}|p{2.0cm}|p{1.6cm}|p{1.9cm}|p{2.0cm}|p{2.0cm}|}

\caption{Comparison of Existing Surveys. The table presents a comparison of existing surveys with respect to their coverage of data validation and authorisation stages.} \label{survey_compirsion} \\
\hline 
\textbf{ } & \textbf{} & \textbf{} & \textbf{} & \textbf{} & \textbf{} \\
\textbf{Survey} & \textbf{Focus} & \textbf{Validation} & \textbf{Validation Stage} & \textbf{Author-isation} & \textbf{Authori-sation Stage} \\
\hline

    \endfirsthead

    \multicolumn{6}{c}{(Continued)} \\
    \hline
    \textbf{Survey} & \textbf{Focus} & \textbf{Validation} & \textbf{Validation Stage} & \textbf{Author-isation} & \textbf{Authori-sation Stage} \\
    \hline
    \endhead

    \cite{9762777} & Security and privacy in AVs & \xmark & Not discussed & Partly & Data transmission, data storage \\
    \hline
    \cite{alam2024data} & Privacy and security of Autonomous Connected Vehicles & Partly & Data exchange and processing & Partly & Data exchange and communication \\
    \hline
    \cite{pali2024autonomous} & Data security in autonomous driving & \xmark & Not covered & Partly & Communication \\
    \hline
    \cite{xu2024data} & Data security in autonomous driving & \xmark & Not covered & Partly & Data collection and exchanging \\
    \hline
    \cite{concas2021validation} & Validation frameworks for AVs & Partly & System-level validation, testing & \xmark & Not covered \\
    \hline
    \cite{omeiza2021explanations} & Explanations in automated driving systems & Partly & Model validation, decision assessment & \xmark & Not covered \\
    \hline
    \cite{9310181} & Software V\&V in AVs & Partly & Training and testing & \xmark & Not covered \\
    \hline
    This Work & Data validation and authorisation in AVs & \checkmark & Covers all stages & \checkmark & Covers all stages \\
    \hline
    
\end{longtable}


%% file: Section2_Background.tex
\section{Background} \label{background}

This section provides a comprehensive overview of the technological foundation of ADS in AVs, including the core components, communication frameworks, and main definitions. It provides foundational definitions to frame the current landscape of autonomous vehicle (AV) technologies.

\subsection{Automated Driving System (ADS)}

\begin{sidewaysfigure}
\centering
\includegraphics[width=1.0\textwidth, height=0.9\textheight, keepaspectratio]{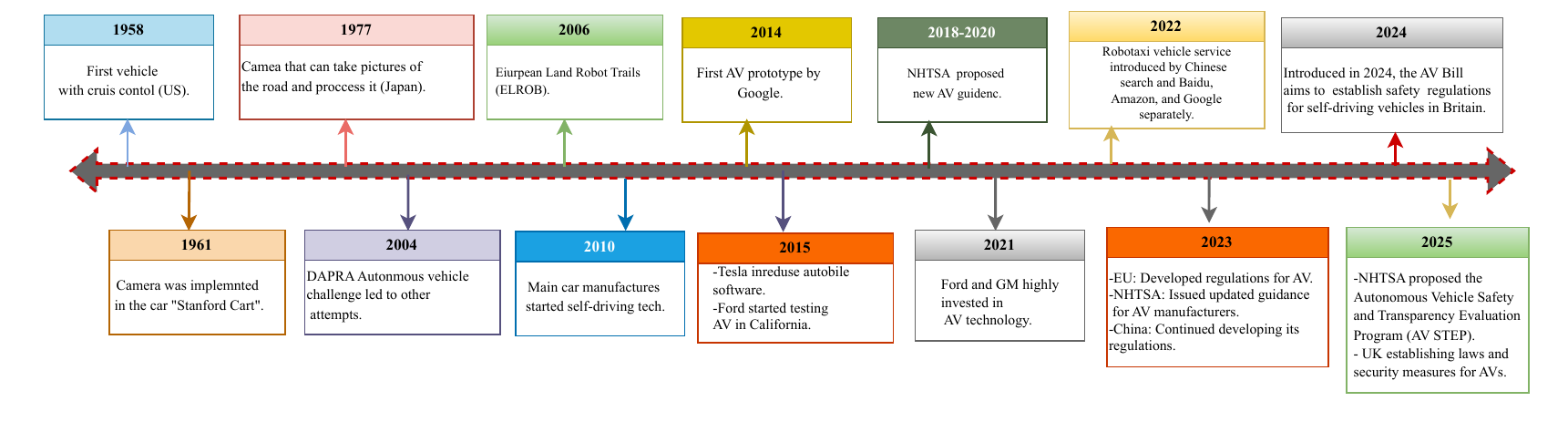}
\caption{\textbf{The timeline of advancement of the autonomous vehicles in the last few decades.} This timeline highlights the ongoing development of safety regulations and standards for AVs in various countries.}
\label{history}
\end{sidewaysfigure}

Continuous innovations have shaped the development of autonomous vehicles (AVs). Figure \ref{history} illustrates the evolution of AVs, highlighting key technological and regulatory milestones from the 1950s to 2025. It begins with the introduction of cruise control in 1958 and includes key milestones such as the integration of road-recognition cameras in 1977.

Driving Assistance Systems (DAS) have played a foundational role in the evolution of AVs, providing critical support for vehicle control and safety. The first generation of DAS utilised sensors that assessed a vehicle's internal condition, primarily focused on safety and stability. In the 1980s, DAS included systems like Traction Control Systems (TCS) and Anti-lock Braking Systems (ABS), which aimed to improve dynamic vehicle stability \cite{rajamani2011vehicle}. In 1995, Electronic Stability Control (ESC) was introduced to further enhance vehicle stability. The second generation of DAS, emerging in the early 1990s, sensors were classified as exteroceptive sensors (e.g., RADAR, LiDAR, and cameras) to detect the external driving environment \cite{pomerleau2012neural}.
The most recent generation of DAS, Advanced Driver Assist Systems (ADAS), has been developed to prevent collisions. A major turning point came in 2004 with the DARPA Grand Challenge, which accelerated AV research and development \cite{thrun2007stanley}. Subsequently, In 2006, the European Land-Robot Trials (ELROB) continued this movement by showcasing semi-autonomous vehicle capabilities. Major car manufacturers began developing automated vehicle technology in 2010. In 2014, Google introduced its first AV prototype \cite{urmson2008autonomous}, followed by Tesla integrated of automobile software for AVs in 2015. In the same year, and Ford commenced AV testing in California \cite{tesla2015autopilot}. Alongside technological progress, regulatory frameworks have rapidly evolved. Between 2018 and 2020, the National Highway Traffic Safety Administration (NHTSA) proposed new AV guidance \cite{nhtsa2020plan}. By 2021, Ford and General Motors had significantly invested in AV technology. Correspondingly, Robotaxi services were introduced by Chinese companies and tech giants like Baidu, Amazon, and Google in 2022 \cite{waymo2022, baidu2022robotaxi}. The year 2023 saw the European Union developing AV regulations, the NHTSA issuing updated guidance for AV manufacturers, and China continuing to expand its regulations \cite{eu2023av, nhtsa2023guidance}. In 2024, the UK introduced the AV Bill to establish safety regulations for AVs, setting the stage for these vehicles to operate on British roads by 2026 \cite{uk2024avbill}. Additionally, In January 2025, the NHTSA proposed the AV Safety and Transparency Evaluation Program (AV STEP). This voluntary program invites vehicle manufacturers, Automated Driving System (ADS) developers, fleet operators, and system integrators to submit detailed information about their AV \cite{nhtsa2025avstep}. These developments reflect the global momentum toward integrating AVs into transportation systems. While an autonomous vehicle (AV) refers to the entire system, including body, mechanical controls, and user interfaces, the Automated Driving System (ADS) denotes explicitly the hardware and software responsible for performing dynamic driving tasks. For clarity, this paper refers to the ADS when discussing the technical system enabling vehicle autonomy.

\subsection{AV Definition}
Autonomous refers to a system’s ability to change its behaviour in response to unanticipated events during operation~\cite{watson2005autonomous}. According to NHTSA \cite{thorn2018framework}, an autonomous vehicle is one as which at least aspects of a safety-critical control function (e.g., steering, throttle, or braking) occur without direct driver input. However, vehicles that provide safety warnings to drivers (for example, forward crash signs) but do not perform a control function are not considered automated.

In September 2018, the NHTSA performed an extensive literature review of all the generic AV system features to identify the attributes that define the operational design domain (ODD). The comprehensive review resulted in 24 ADS features, specifically describing functionality and proposed timelines for commercial deployment across the different Society of Automotive Engineers (SAE) International levels of driving automation. Accordingly, the SAE's six-level taxonomy has become the widespread industry standard\cite{sae2021sae}:

\begin{itemize} 
\item[] Level 0: No Automation – The human driver performs all driving tasks, and any system support (like warning systems) does not automate driving. \item[] Level 1: Driver Assistance – The vehicle may assist with a specific task, such as steering or acceleration, but the driver must remain in control and perform all remaining aspects of driving. \item[] Level 2: Partial Driving Automation – The vehicle can control both steering and acceleration/deceleration, but the driver is responsible for monitoring the environment and must remain engaged at all times. \item[] Level 3: Conditional Driving Automation – The vehicle can handle all aspects of driving in specific conditions or environments, but the driver must be prepared to take over when requested. \item[] Level 4: High Driving Automation – The vehicle is capable of performing all driving tasks in specific conditions (such as certain road types or geofenced areas), and driver intervention is not required, though manual control is possible. \item[] Level 5: Full Driving Automation – The vehicle performs all driving tasks in all conditions and environments, with no need for driver intervention at any time. 
\end{itemize}

\subsection{AV Architecture}

Researchers studying AV focus on two main areas:  defining their components and understanding their functional perspective. Some papers focus on the technical aspects of AV components, such as in \cite{betz2022autonomous}, while others take a functional approach~\cite{sharma2021recent, yeong2021sensor}. From a technical perspective, AVs have two main layers: hardware and software. Each layer is comprised of several subcomponents. There is some disagreement among researchers about categorising the core competencies of different subsystems when defining the functional perspective of AVs.

\begin{figure}[H]
\centering
\includegraphics[width=3.5 in]{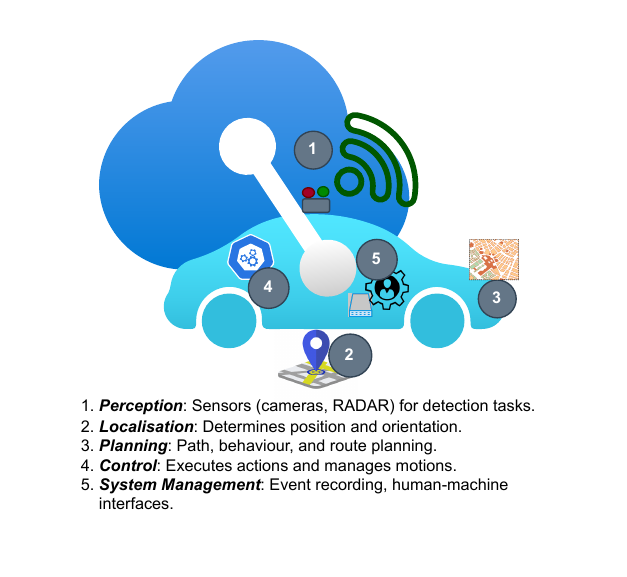}
\caption{\textbf{Autonomous Vehicle (AV) Main Operations} This diagram illustrates the main operations of an AV. It highlights five key areas: (1) Perception, where sensors such as cameras and RADAR collect data and detect objects; (2) Localisation, which determines the vehicle's position and orientation; (3) Planning, involving path, behaviour, and route planning; (4) Control, which executes the planned actions and manages the vehicle's motions; and (5) System Management, responsible for event recording and human-machine interfaces}
\label{operation}
\end{figure}

 However, in general, AV systems are made up of three to five primary functions: perception, localisation, planning, control and navigation, and system management~\cite{omeiza2021explanations,vargas2021overview}, as illustrated in Figure \ref{operation}.

In the Figure, the several AV fundamental operations are as follows:

\begin{enumerate} 
  \item \textbf{Perception} refers to collecting data and extracting relevant understanding from the environment, such as the detection of road signs, as well as object detection and classification~\cite{pendleton2017perception, reid2019localization}. These detection tasks are performed by various sensors such as cameras and Radio Detection And Ranging (RADAR).

 \item \textbf{Localisation} refers to the ability of the AV system to determine the vehicle’s position and orientation relative to the environment.
\item \textbf{Planning}
      consists of three stages: 
\begin{itemize} 
\item[-] Using algorithms, the path planner calculates the most efficient geometric path.
\item[-] The behaviour planner determines the optimal behaviour based on the path planned by the path planner.
\item[-] The estimation of the best possible route subject to vehicle dynamics and environmental constraints~\cite{sharma2021recent}.
\end{itemize}
\item \textbf{Control} refers to executing planned actions and managing the vehicle's motions, such as changing lanes~\cite{omeiza2021explanations,pendleton2017perception}. 
\item \textbf{System management} includes all the functions related to event data recorders, human-machine interactions involving in-vehicle interfaces~\cite{omeiza2021explanations}, and external human-machine interfaces.      
\end{enumerate}

An alternative architectural approach, such as in \cite{badue2021self}, divides AV systems into two primary components: the perception system and the decision-making system. While the perception system is further broken down into subsystems in charge of tasks like localisation, static obstacle mapping, and moving obstacle detection, the decision-making system is segmented into tasks like route planning, path planning, behaviour selection, and motion planning. This approach organises the autonomy system of the AV by highlighting discrete levels of perception and decision-making.
Architectural paradigms like end-to-end and layered architecture further classify AV system designs\cite{zhao2023autonomous}. The layered architecture divides the system into perception, including simultaneous localisation and mapping (SLAM), planning, and control layers, ensuring a modular and structured approach. While the end-to-end architecture processes raw sensor data directly to output control commands using deep learning techniques, providing reduced
system complexity. 

{\textit{\textbf{Sensors and Sensors Fusion}}}
\\
For an AV vehicle to have a high-quality and real-time understanding of its surrounding environment, it must quickly and accurately detect, comprehend, and track all objects. As a result, relying on a single source to generate all necessary data is impossible. Multiple sensors are equipped to provide both perceptual and location views of the environment, allowing the vehicle to make real-time decisions. Sensors are devices that translate detected objects or changes in the surrounding environment into quantitative measurements for processing~\cite{yeong2021sensor}.\newline
{\textit{\textbf{Sensors Types}}} \\\\
AV systems, while having some variations, all consist of two primary types of sensors based on their operational principle: Exteroceptive and Proprioceptive. Exteroceptive sensors are external state sensors that are utilised to perceive the environment, such as calculating the distance to objects or light intensity from the surroundings of the system. Examples of these sensors include cameras, RADAR, etcetera. Proprioceptive sensors, on the other hand, are internal state sensors that capture the dynamic state and measure the internal values of a dynamic system. Examples of this type of sensor include Global Positioning Systems (GPS), encoders, accelerometers, etc. \cite{yeong2021sensor,fayyad2020deep}. The most commonly used sensors for AV are listed in Table \ref{tab:common_sensors}. In addition, Ahangar et al. \cite{ahangar2021survey} provided a detailed comparison of sensors and their individual challenges.
As each sensor has different advantages and limitations in other aspects, integrating sensors is necessary to obtain an optimal perspective, and this operation is known as "sensor fusion". \\\newline

\textbf{\textit{Sensor Fusion}}\\\\
The advantage of this operation is to combine the data originating from different sources to complete each other's functions to provide an improved outcome in some specific criteria and data aspects for decision tasks~\cite{nguyen2021fusing}. The best example to explain the benefit of this method is to fuse RADAR sensors and camera images, where these sensors have different strengths and weaknesses. RADAR is not affected by the illumination of the environment, but it is not able to provide accurate data regarding an object's body. On the other hand, the Camera's image could provide these data, however, it may provide conflicting data under some illumination conditions~\cite{campbell2018sensor}. The initial step in data fusion is sensor calibration, which means notifying the ADS regarding the sensors' position and orientation to collect and associate the data in space and time~\cite{wishart2020literature}. While the various sensors capture data in relation to the same object, the produced data could be combined to obtain different information to achieve higher-quality output. Other explicit examples with details and fusion techniques are discussed by Campbell et al.~\cite{campbell2018sensor} and Fayyad et al.~\cite{fayyad2020deep}.  \\

\begin{table*}[!htb]
\caption{\textbf{Most Commonly Used Sensors.} The table lists the most commonly used sensors in AVs, categorised into exteroceptive and proprioceptive sensors. Exteroceptive sensors gather information from the external environment, while proprioceptive sensors provide information about the vehicle's internal state.}
\label{tab:common_sensors}
\centering
\resizebox{\textwidth}{!}{%
\begin{tabular}{lp{12cm}} 
\toprule
\multicolumn{2}{c}{\textbf{Exteroceptive Sensors}} \\ \midrule
\textbf{Sensor}  & \textbf{Function} \\ \midrule
Light Detection and Ranging (LiDAR ) & Uses light beams to provide a 360-degree distance between an object and the car. \\  
Radio Detection and Ranging (RADAR) & Uses radio waves to determine the distance between an object and the car. \\ 
Camera & provides images of the environment to interpret data. \\ 
Ultrasonic & Used for short-distance object detection, such as parking. \\ \midrule
\multicolumn{2}{c}{\textbf{Proprioceptive Sensors}} \\ \midrule
GPS (Global Positioning System) & Locates the vehicle. \\ 
IMU (Inertial Measurement Unit) & Measures acceleration and angular rate. \\ 
Encoders & Provide feedback signals used in speed and/or position control. \\ 
Accelerometers & Measure acceleration. \\ \bottomrule
\end{tabular}%
}
\end{table*}

\subsection{Communication in Autonomous Vehicles}
AVs are able to communicate with any compatible systems, including other AVs, infrastructure, and pedestrians. This communication is known as Vehicle to Everything (V2X) technology, referring to how the vehicle communicates with everything. The network must be continuously fast, reliable, and secure to achieve an efficient cooperative environment with minimal delay (latency). Specifically, there are two main tendencies used all over the world:  the wireless standard 802.11p or mobile networks, especially 5G \cite{martinez2018autonomous, szalay20205g, guleng2020edge}. In the foreseeable future, sixth-generation (6G) wireless systems will be crucial for V2X communications in AV \cite{kulshrestha2024disruptive, noor20226g}.

The differences between conventional and automated vehicles are apparent; however, the concept of connected vehicles represents a distinct phase in automotive technology. While the literature sometimes blurs the distinction between connected vehicles and automated vehicles (AVs), connected technology is a critical step toward achieving full automation. A connected vehicle is part of the Internet of Things (IoT), enabling data exchange, software updates, and communication with other vehicles and infrastructure. In these smart vehicles, all electronic control units (ECUs) and onboard units (OBUs) are interconnected through multiple digital buses, such as the Controller Area Network (CAN), Ethernet, FlexRay, Local Interconnect Network (LIN), Media Oriented Systems Transport (MOST) and Bluetooth~\cite{fulbright2017privacy, cebe2018block4forensic}. 
Such capabilities with sensing, communicating with the surroundings, and controlling the driving tasks represent the AV. The Connected and Automated Vehicle (CAV) is the vehicle that performs automated driving tasks and connectivity with other vehicles, road users, the road infrastructure, and the cloud \cite{guanetti2018control}. \\ 
\indent While conventional cars constitute the vast majority on today's roads, forecasts suggest a notable rise in the presence of connected and autonomous vehicles in the coming years. Expectations indicate that the number of connected cars will reach 700 million by 2030, while the number of AVs will exceed 90 million \cite{fulbright2017privacy}. Correspondingly, Zhang et al.~\cite{zhang2019impact} address the problem of optimally controlling CAVs under mixed traffic conditions where both CAVs and conventional vehicles are together on the roads. Other studies \cite{yi2024access, ying2024infrastructure} subsequently aim to address the challenges and opportunities that arise and to leverage the capabilities of CAVs to enhance traffic flow at unsignalised intersections while ensuring safety in a mixed traffic environment where both conventional vehicles and connected and automated vehicles (CAVs) coexist on the roads.

\subsubsection{Communication Technologies}
Communication technologies play a critical role in enabling AV to interact with their environment, improving safety, efficiency, and driving experience. In this section, we explore several key communication technologies that are vital for vehicular networking, including Vehicular Ad-Hoc Networks (VANETs), and their associated communication types such as Vehicle-to-Vehicle (V2V), Vehicle-to-Infrastructure (V2I), and Vehicle-to-Everything (V2X).

    \textbf{ Vehicular Ad-Hoc Networks (VANETs):}\newline
    A group of vehicles connected via a wireless network \cite{marinescu2022cloud}. This network is a basic part of the Intelligent Transportation System (ITS) framework to assist various connections:
    \begin{enumerate}
        \item Vehicle-to-vehicle (V2V) communication that allows the vehicles to communicate with each other and share the necessary information, such as traffic jams. To establish V2V communication, vehicles should have an On Board Unit (OBU), Omnidirectional antennas, sensors and actuators, and a Global Positioning System (GPS)~\cite{sharma2019vehicular}. 
        \item Vehicle-to-Infrastructure (V2I) communication enables the vehicles to interact with the roadside units RSUs which are fixed devices installed next to the road covering a dedicated area. These communications are mostly conducted by using wireless dedicated short-range communications (DSRC), which aims to provide active safety and convenience services \cite{guanetti2018control}. 
        \item Vehicle-To-Everything (V2X) Communication allows the vehicle to communicate with other entities using technologies such as Cellular V2X (C-V2X), including 5G and 6G.  
    \end{enumerate}

\begin{figure}[!t]
\centering
\includegraphics[width=3.5in]{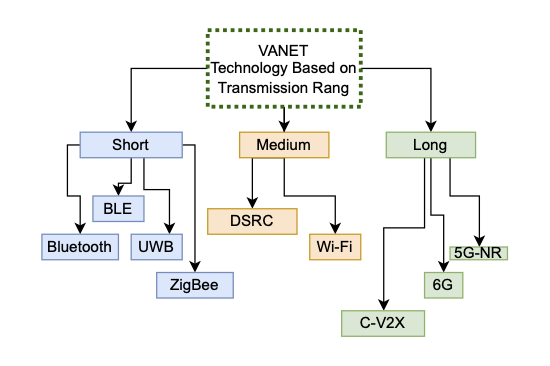}
\caption{VANET technologies based on transmission Range \cite{ahangar2021survey}.}
\label{Different VANET technologies based on transmission Range}

\end{figure}

VANETs networks utilise different communication technologies based on their transmission range, which can be categorised into short, medium, and long-range communication as shown in Figure~\ref{Different VANET technologies based on transmission Range}.
\begin{enumerate}

    \item \textbf{Short-range communication} includes Bluetooth, Ultra-Wideband (UWB), and ZigBee, which are wireless communication technologies used for vehicle-to-vehicle (V2V) communication in dense traffic environments due to their low power consumption and short operational range.  Bluetooth, especially Bluetooth 5, the latest version,  offers low-cost, low-energy communication up to 200 m, but faces interference and delays in dense settings. In contrast, UWB provides robust, energy-efficient communication with strong obstacle penetration and resistance to multipath fading, ideal for non-line-of-sight environments. Relative to other wireless protocols, ZigBee is simple, energy-efficient, and supports self-healing multi-node networks, though it may suffer interference on shared channels~\cite{stefanovic2024overview}.
    \item \textbf{Medium-range communication} relies on Dedicated Short-Range Communication (DSRC) and Wi-Fi, which provide higher data rates for V2V and V2I applications.
    \item \textbf{Long-range communication} involves Cellular-V2X (C-V2X), 5G NR, and emerging 6G technologies, which support high-speed, low-latency communication for advanced AV applications.
    
\end{enumerate}
Ahangar et al. presented a comprehensive survey  regarding different vehicle communication technologies,  their application, limitations, and advantages ~\cite{ahangar2021survey}.\\
5G is suitable for long-range transmission to meet the high-mobility demand architecture; while 6G is still in its conceptualisation phase, although the technology-driven key performance indicates extremely serves the AV's communication requirements~\cite{he20206g}. In general, various regions prefer DSRC over C-V2X (e.g., USA) due to the heavily deployed infrastructure around the country, however, this recently changed when Europe decided to move with cellular-based technology for CAVs~\cite{khan2022level}.

 In Brussels (2018), the 5G Infrastructure Public-Private Partnership (5G PPP), which is a cooperative combined initiative between the European Commission and the European ICT industry (ICT manufacturers, telecommunications operators, service providers, SMEs, and researcher Institutions,  launched to deliver solutions, architectures,  and standards for the next generation communication infrastructures \cite{5gppp}. The survey provided by Hakak et al. \cite{hakak2022autonomous} highlights and summarises the key projects related to the 5G AV.
\subsection{Autonomous Vehicle Potential Benefits and Costs}\label{Benefits}
Automation technology is a promising future for safety, mobility, environment, and luxury \cite{srivastava2022autonomous}. It may reduce crash risks by avoiding human error and distracted driving. Many conventional vehicle crashes occur due to human error and distracted driving. In the USA, partially automated crash avoidance features could reduce the severity of as many as 1.3  million crashes every year, including 133,000 injury crashes and 10,100 fatal crashes~\cite{harper2016cost}. Thus, traffic will disregard poor human driving behaviour and improve performance in terms of road safety and traffic congestion. In addition, AV contributes significantly to reducing emissions (worldwide goal to achieve net-zero by 2050~\cite{netzero}).\\ Comparatively, even though adopting automation will increase safety, there are specific safety concerns related to its development. Its total or partial dependence on driving assistance systems results in a serious risk with both hardware and software issues. Additionally, sensors may be compromised due to environmental conditions such as dangerous weather. Another significant concern is that through cyber attacks, an automobile and/or its technological environment may be subject to causing grave privacy and security issues. Other benefits and costs are summarised in Table ~\ref{Pros_Cons}.\\

\subsection{Data Authorisation and Validation in Autonomous Vehicles Ecosystem}
Data plays a critical role in AV. Data is constantly being created, exchanged, and stored in this dynamic environment, creating challenges in data validation and authorisation that have not received the needed attention \cite{sun2021survey}.

\begin{table} [htbp] 
\centering
\caption{Pros and Cons of applying Autonomous Vehicles.}
\label{Pros_Cons}

\scalebox{0.75}

\begin{tabular} {|l|l|} 
 \hline 
 \textbf{Pros}  & \textbf{Cons} \\[1.0ex]
 \hline
\parbox[t]{5cm}{\textbf{\\Increased safety:}\\
Accidents will be greatly avoided due to the various assistance systems, ongoing connection, and connectivity between vehicles.\\}  
 
 & 
\parbox[t]{5cm}{\textbf{ \\Increased infrastructure costs:} \\ AVs require higher standards for road maintenance and design.\\} 
 
\\ \hline

 \parbox[t]{5cm}{\textbf{\\Reduced energy consumption and pollution:} \\ Since these vehicles are supposed to run on sustainable energy, carbon and emissions of greenhouse gases will be almost nonexistent.\\}
  &
  
 \parbox[t]{5cm}{ \textbf{\\System failure risks:} \\
Hardware/software failures, wrong data, feeding/processing errors, faster traffic speeds, and increased overall vehicle travel are additional collision causes that may be on the rise.\\}  

 \\ \hline

 \parbox[t]{5cm}{\textbf{\\Reduce traffic congestion:}\\
Although AVs move at a slower speed in cities, the traffic efficiency will be higher because of the efficient connection between vehicles.\\} 

&

\parbox[t]{4cm}{\textbf{\\Data protection issues:} \\ 
The network's environment causes security and privacy issues.} 
\\ \hline
 
\end{tabular}

\end{table}

\paragraph{Data Authorisation}
Data authorisation in the AV field refers to the set of mechanisms and policies that determine \textit{who} can access specific data within an AV system, at \textit{which} stage of the data lifecycle, under \textit{what} conditions and \textit{how} this access occurs. 
These mechanisms mainly draw from well-established models such as attribute-based access control (ABAC) \cite{hu2013guide}, role-based access control (RBAC) \cite{sandhu1998role}, and usage control (UCON) \cite{park2004uconabc}.

For instance, during a sensor maintenance operation, only authorised suppliers can access vehicle performance data for maintenance purposes in the required stage. Similarly, passengers' information, such as location history, may only be accessible to authorised parties (e.g., regulatory bodies) during an investigation, while remaining restricted at other stages to protect privacy.
\paragraph{Data Validation}
Data validation in AV ensures that data used by the AV ecosystem at any stage of the data lifecycle is accurate, consistent, and reliable. This process includes sensor fusion,  cross-verification, and cryptographic mechanisms to ensure trustworthiness \cite{christidis2016blockchains, yeong2021sensor}. 
For instance, the system validates data from multiple sensors to confirm the detection of an object. Another example is validating accident-related data to ensure its accuracy and integrity.  
\paragraph{Authorisation and Validation Requirements for AV Data}
Based on inspiration from the literature, such as \cite{wolter2010approach, khan2022authorization}, this work defines the Authorisation and Validation Requirements for AV as:

\textbf{Requirement 1: }
Access permissions must be distributed among multiple stakeholder roles (e.g., manufacturer, supplier, user, regulator), each with clearly defined access rights. 

\textbf{Requirement 2: }
These access control rules and mechanisms should be adaptable, allowing for any changes in access rights, and customisable to ensure flexibility as the system grows and evolves.

\textbf{Requirement 3: }
Stakeholders who generate data (e.g., passengers, owners, or manufacturers) must not have unrestricted access to all aspects of the data.

\textbf{Requirement 4: }
Any critical access actions must involve approval from multiple independent stakeholders to ensure accountability.

\textbf{Requirement 5: }
All these critical access actions must be recorded and traceable through secure logging mechanisms that are only accessible to authorised entities.

\textbf{Requirement 6: } 
Data validation processes must be conducted by domain experts, such as cybersecurity analysts and accident detectives to detect any unauthorised access attempts and ensure data integrity.

\textbf{Requirement 7: } 
Incorporated security measures must be applied as the data generated by external sources (e.g. V2X communication) may contain inaccuracies or malicious inputs.

\textbf{Requirement 8: }  
Data collected by the vehicle from the environment (e.g. pedestrians, other vehicles) may contain third-party information that the entity authorised to access the vehicle's data is not authorised to view or process. Mechanisms must ensure that such data is filtered or anonymised to respect the privacy and rights of external parties.

\textbf{Requirement 9: }
Critical or personal AV data must not be leaked in any way to external, unauthorised entities (e.g. service providers or insurance companies) without explicit agreements.

%% file: Section3_dataflow.tex
\section{Data Flow in Autonomous Vehicle Systems: Stages, Challenges, and Solutions}\label{Dataflow}

\begin{figure*}[htbp]
  \centering
  \includegraphics[width=5.0in]{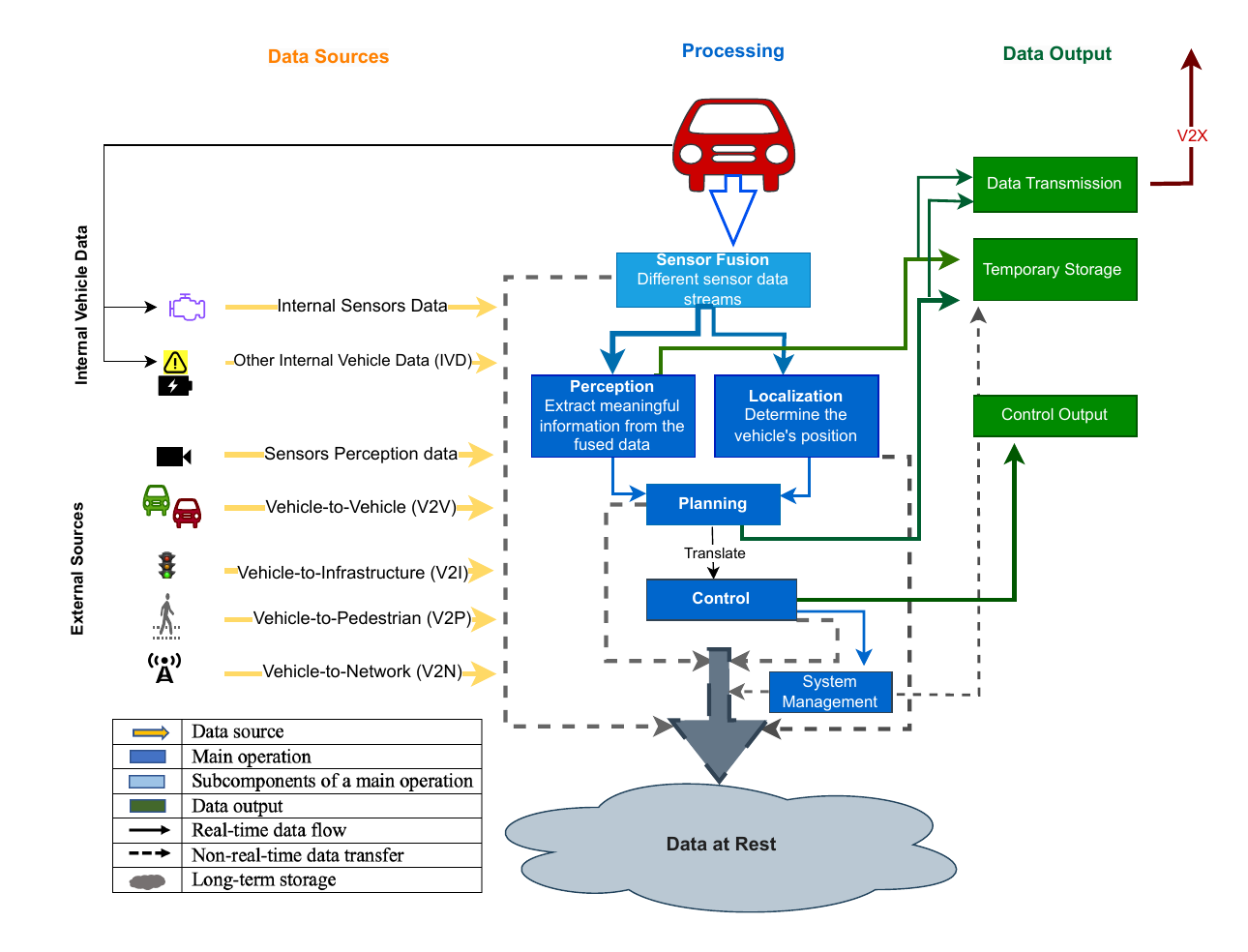}
  \caption{\textbf{Data Flow in an Autonomous Vehicle} This diagram depicts the data flow in an AV system. It shows how sensor data and external sources are fused and processed through the Perception, Localisation, Planning, and Control stages. The outputs include Control commands, Data Transmission (V2X), and Temporary Storage, with all processes contributing to long-term Data at rest storage. }
  \label{Dataflowfig}
\end{figure*}

The operation of AV depends on a continuous and efficient flow of data. As illustrated in Figure ~\ref{Dataflowfig}, this data originates from various sources such as sensors and external environments, undergoes multiple layers of processing at the local and edge levels, and eventually is stored or analysed in backend systems. Each of these stages presents challenges that are unique, common, and intersecting and have been discussed in this section along with current solutions. 

\subsection{Data Sources}
The vehicle's hardware and software systems collect and process substantial data to operate safely. These data sources might be categorised as sensory observation, Intelligent Transportation System (ITS), Geographic Information System (GIS) and map-related sources, social media feeds of drivers or passengers, linked data, archive, and legacy data \cite{arooj2022big}.   As shown by yellow arrows in Figure \ref{Dataflowfig},  these data sources may alternatively be categorised as:

 \begin{itemize}
     \dashitem \textbf{ Internal Data Sources}, which are collected and processed within the vehicle itself, include:

\begin{enumerate}
\item Internal sensor data: These sensors monitor various parameters and parts of the vehicle’s internal systems, such as engine and brake sensors.

\item Other internal vehicle data (IVD): This includes other data, such as the data about the owner and data that comes from the vehicle's status; for example, if the tank or battery is almost empty. 
 \end{enumerate}
 \dashitem \textbf{External Data Sources} are obtained from sources outside of the vehicle itself, which are crucial for various aspects of driving, navigation, and safety. In the same Figure \ref{Dataflowfig} these sources include:
 \begin{enumerate}
 
\item Vehicle-to-Vehicle (V2V) represents sharing real-time information about other cars' speed, direction, and braking status. 

\item Vehicle-to-Pedestrian (V2P) communication includes exchanging information between vehicles and pedestrians or other detected road users, such as cyclists.

\item Vehicle-to-Infrastructure (V2I) refers to the exchanged information between vehicles and roadside infrastructure or traffic management systems.

\item Vehicle-to-Network (V2N)
represents the connection between the vehicle and the cloud network to access navigation services, real-time traffic updates,  weather, and entertainment content.

\item Vehicle-to-Everything (V2X) refers to the communications between vehicles and other entities, including cars (V2V), pedestrians (V2P), infrastructure (V2I), and vehicle-to-network (V2N). In the figure, V2X represents the comprehensive output integrating processed data from V2V, V2P, V2I, and potentially other sources, facilitating interaction with the broader environment, including vehicles, pedestrians, and infrastructure.

	\end{enumerate}

 \end{itemize}

\paragraph{Key Challenges}
\begin{enumerate}
    \item Volume and Variety: The vast amount of data generated by AV poses significant challenges at every stage of its lifecycle. 
For instance, cameras produce 20–40 MB of data per second, while Light LiDAR systems generate between 10–70 MB per second \cite{liu2020computing}. The overall data volume and variety become extreme challenges with numerous heterogeneous devices and diverse communication streams. In addition to this wide variety of types and formats, the AVS must process in real-time to support effective driving decision-making. Another challenge is the authorisation to access and use data from these multiple heterogeneous devices and systems, especially when integrating third-party sensors or external communication streams. Since these sources often rely on data from other devices, ensuring secure and controlled access and sharing is crucial to maintaining trust and safety in the AV ecosystem.

\item Data Quality and Reliability: 
Robust data validation is vital to maintain data integrity, especially in such environments with a high potential for noisy or incomplete data. In addition, the flow fusion information from multiple sources must identify and filter out corrupted or redundant data. Without adequate data validation and fusion, invalid or incorrect inputs could affect decision-making, potentially leading to unsafe driving behaviours. 
\end{enumerate}
\paragraph{Current Solution}
Many recent studies have sought to ensure reliable data management, integrity, and efficient communication among sources at this stage. This includes data exchanged between vehicles, RSUs, stations, and data that reaches the sensors inside the vehicle. Some research suggests that blockchain technology could effectively address these challenges and meet data management requirements at this stage. For example, a blockchain model has been proposed in \cite{10444480} that utilises the Hashgraph consensus algorithm to facilitate decentralised and secure data sharing among nodes. Each node disseminates information through a gossip protocol, enabling rapid consensus and verification of data integrity. Furthermore, instead of data encryption, Changvala et al. proposed a method to hide the integrity of LIDAR and RADAR data. \cite{changalvala2021sensor}. Similarly, JAVED et al. \cite{javed2020odpv} proposed a protocol to isolate false data in V2X communications messages. Another data integrity verification scheme is proposed in \cite{shen2022innovative}; it aligns GPS data with other information regarding passengers to make sense of the vehicle’s reliability. In addition, leveraging RSUs allows this data integration to be used for integrity checks. Another approach \cite{Rakhmanov2023GNSS} focuses on a hybrid GNSS data compression method for autonomous vehicles, enhancing data transmission efficiency and resilience through frame differencing and entropy coding. Although its applicability may be limited, it achieves a good compression ratio and maintains reliability.

Additionally, in a notable attempt to improve the quality of sensors' data, Min et al. \cite{min2023fault} have proposed a framework with two methods: first, a residual consistency checking algorithm that utilises sensor redundancy to isolate faulty sensors, and second, a Denoising Shrinkage Autoencoder (DSAE) that enhances anomaly detection in sensor data. Despite the algorithm's inability to isolate the "Spike" anomalies due to their brief duration, these methods help ensure that the sensors in AV are working correctly and provide reliable data. Similarly, to ensure that only trustworthy data is used for decision-making in AV operations, a study has integrated data quality metrics with a trust and reputation model\cite{9174269}. This mechanism evaluates the correctness and reliability of real-time data sources based on their past behavior and interactions. While these solutions improve authorisation and data validation for AV data sources, further research is still needed in these areas and to address challenges like interoperability in data resources.


\begin{table}[!t]
\caption{Sample Event Data Recorder (EDR) Parameters Captured During a Vehicle Event~\cite{tesla} }
\label{edr_parameters_example}
\centering
\begin{tabular}{|l|l|}

\hline
\textbf{Data Event}  & \textbf{Value}                              
\\ \hline
Maximum Delta-V. Longitudinal (km/h) & -61
\\ \hline   
 Time To Maximum Delta-V. Longitudinal(ms) & 95.0
\\ \hline

Maximum Delta-V Lateral (km/h) & -1
 \\ \hline
Time To Maximum Delta-V Lateral (ms) &  72.5
\\ \hline
  Time To Maximum Delta-V Resultant (ms) & 95.0
  \\ \hline
  Ignition Cycle At Event & 271
 \\ \hline
 Ignition Cycle Runtime (minutes)&  10.3
 \\ \hline
 Odometer At Event Time Zero (km)  & 30.5
  \\ \hline
  Airbag Warning Lamp Status & Off
  \\ \hline
  ABS Warning Indicator Status & Off
  \\ \hline
  Vehicle Drive Mode & Natural
  \\ \hline 
  Driver Safety Belt Status & Buckled
  \\ \hline
  Passenger Safety Belt Status & Buckled
  \\ \hline
  Occupant Classification Status in Front Passenger Seat & Small Adult
  \\ \hline
  Driver Seat Track  Position & Rearward
  \\ \hline 
  2nd Row Left Safety Belt Status & Not Buckled 
  \\ \hline
  2nd Row Left Seat Occupant & Not Occupied 
  \\ \hline
   2nd Row Center Safety Belt Status & Not Buckled
   \\ \hline 
 2nd Row Center Seat Occupant & Not Occupied
   \\ \hline
 2nd Row Right Safety Belt Status & Buckled
 \\ \hline 
 2nd Row Right Seat Occupant & Not Occupied
   \\ \hline 
  3rd Row Left Safety Belt Status & Not Available 
  \\ \hline
  3rd Row Left Seat Occupant & Not Available 
  \\ \hline
  3rd Row Right Safety Belt Status &  Not Available 
 \\ \hline 
     3rd Row Right Seat Occupant & Not Available  
  \\ \hline 
Driver Airbag Deployment 2nd Stage Disposal & Yes
 \\ \hline
 Right Front Passenger Airbag Deployment 2nd Stage Disposal & Yes
     \\ \hline
 Complete File Recorded & Yes
   \\ \hline
   
\end{tabular}

\end{table}


\subsection{Local Data Storing }
Based on sensors and cameras positioned in various regions inside and outside the automobile, the vehicle creates and maintains data. According to how many ignition cycles the car goes through, most of them will hold their data for a while before replacing it with fresh data. Local storage systems, such as sensor data buffering,  communication cache, and artificial intelligence (AI) models of data, play a critical role in managing large volumes of data generated by AVs. However, this section focuses on the event local recorder tools. Besides the data sent to the backend servers and cloud, an Event Data Recorder (EDR) and Data Storage System for Automated Driving (DSSAD) are the primary tools for storing significant event data. The rest of this section focuses on these technologies, their functions, gaps, and current solutions.

\subsubsection{Evolution of Transportation Data Recorders}

The first practical transportation data recorder was introduced in 1921. It records vehicle speed, engine RPM, and distance moved onto a rotating circular chart \cite{correia2001utilizing}. 

\begin{sidewaysfigure}
\centering
\includegraphics[width=1.0\textwidth, height=0.9\textheight, keepaspectratio]{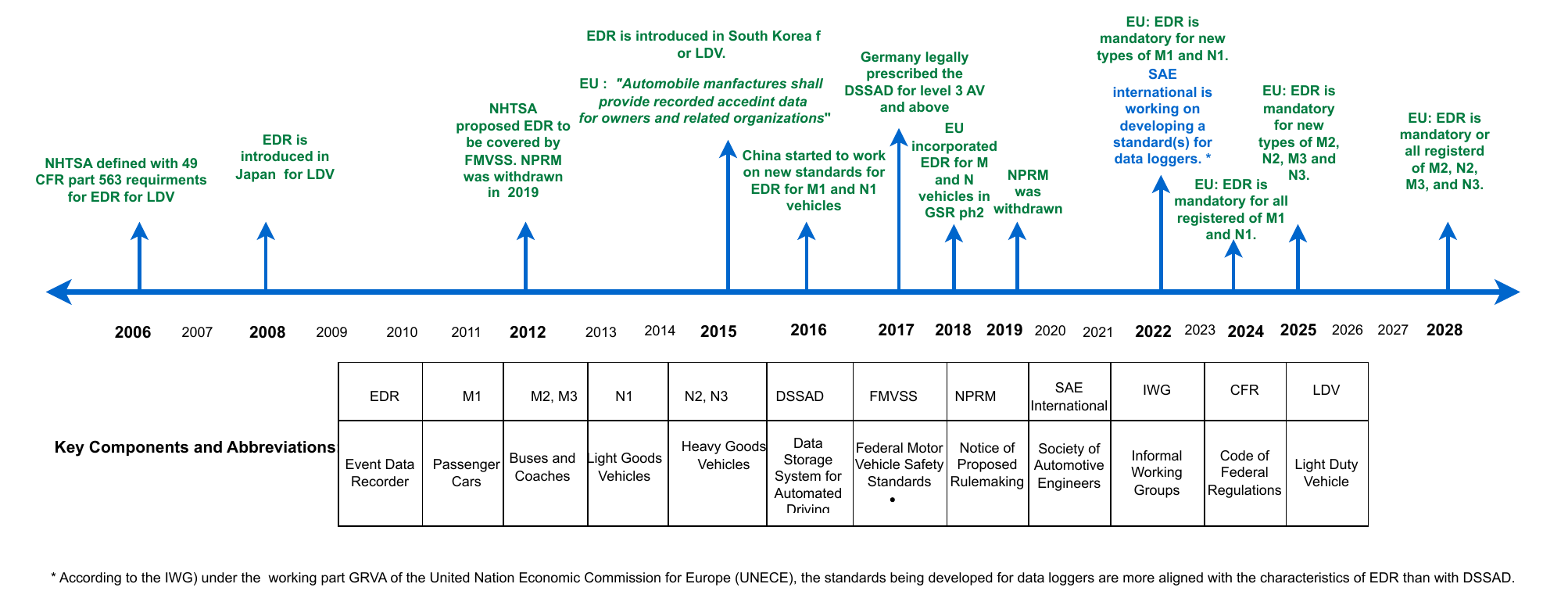}  
  \caption{\textbf{Timeline of EDR and DSSAD Regulations and Standards Development (2006–2028)} \cite{Oica, bohm2020new, government}. This figure illustrates key events in the development and implementation of regulations and standards for EDR and DSSAD across various regions.}
  \label{existingactivities}

\end{sidewaysfigure}

Figure~\ref{existingactivities} shows the current development movement of the existing national and regional activities on this technology. The diagram outlines the global timeline of the adoption of the EDR. While South Korea and Japan were early adopters, the EU mandated EDRs for new and registered vehicles, and the US has ongoing discussions about implementing them. International organisations like SAE and IWG are working on developing standards for data loggers, including EDRs\\

\subsubsection{\textbf{EDR}}
 Event Data Recorder (EDR) is a recording device inside the vehicle that can capture information regarding an event  \cite{correia2001utilizing} in a readily usable manner. Its concept is the same for both AV and conventional vehicles; there is a list of all vehicles with EDR in \cite{edrlist}. Such devices would support real crash investigations and analyses of performance due to their data capturing of a few seconds, both pre- and post-crash. Useful and meaningful data, such as vehicle speed and brake status, are stored. For example, the Tesla Model 3 (Autopilot) is designed to record data such as how various vehicle systems were operating, whether or not driver and passenger safety belts were fastened; how far (if at all) the driver was depressing the accelerator and/or brake pedal; and, how fast the vehicle was traveling \cite{tesla}. Table~\ref{edr_parameters_example} is an example of such recorded data in detail. The technique behind this device generally stores data with the airbag control module once deployed. Other recording techniques and event data recorders have been explained comprehensively in \cite{correia2001utilizing}.

 In December 1996, the National Highway Traffic Safety Administration (NHTSA) and the National Aeronautics and Space Administration (NASA) cooperated to improve airbag safety. Consequently, NASA recommended that NHTSA encourage installing and obtaining crash data for safety analyses from vehicle crash recorders \cite{correia2001utilizing}. It is worth indicating, however, that (EDR- AV), the EDR for automated vehicles, must considerably exceed the performance of EDR for conventional vehicles (EDR-CV) \cite{gwehenbergerneeds}. Publications worldwide seek to provide “guidelines for the development of automated driving functions”, including identifying the technical requirements for EDR-AV. The  Regulation (EU) 2019/2144  sets rules on technical requirements for new types of motor vehicles to maintain safety and environmental protection  \cite{EUapproval}. According to that regulation,  in June 2022, the EDR is supposed to become mandatory for all vehicles with SAE Level 3/Level 4 functions \cite{gwehenbergerneeds, eu2024}. From July 2024, all newly registered passenger cars in the EU must be equipped with an EDR \cite{eu2024}. In addition, the IWG EDR/DSSAD deals with the definition of the technical requirements as a prerequisite for the corresponding UN regulations for both EDR systems (conventional/autonomous) \cite{connectedeurupe}.\\
All crash data must be stored, available, and retrieved for crash reconstruction purposes, even for testing, to understand the crash circumstances \cite{national2015url}. However, for AV level 3 and above, where the system can perform all driving tasks for a specific time, EDR is not helpful regarding responsibility and liability without the Data Storage System for Automated Driving (DSSAD) system to indicate who was in control during the accident.

\subsubsection{DSSAD}
Data Storage Systems for Automated Driving (DSSAD) is a storage device that determines whether the system or the driver has controlled the vehicle \cite{bohm2020new}. This provides information regarding who controlled the driving task at a particular time to decide the responsibility and, consequently, liability issues. 
Examples of stored data elements include position and time-stamped switches of the ADS from one mode to another \cite{Oica}. These may contain data about whether the system is activated, manually or automatically deactivated. Therefore, it also records time and position and when the driver is requested to drive;
position and time-stamped override through brake control by the driver; 
position and time-stamped transition demand by the ADS. Thus, the nature of the reasons for a transition demand or deactivation can be determined, such as:
\begin{itemize}
    \item Driver not available or lack of driver attention.
    \item Driver override.
    \item System failure.
    \item Unplanned event (ex: bad weather).

\end{itemize}
While EDR is event-based storage, DSSAD is the continuous storage of specific AD data sets. All of these data must be clearly identified and recorded to eliminate confusion or misinterpretation.

\subsubsection{Limitations of Traffic Accident Data Records}

 Despite the important roles that these tools play in supporting investigations, they have the following limitations:
\begin{enumerate}

\item Storage and Data Availability:

Due to storage limitations and the large number of continuously variable parameters required for analysis, EDR stores only specific data parameters for a short time in specific events, such as events with airbag deployment. 
 In addition, if these tools do not have sufficient space to record an event, they will overwrite or erase the previous event data.

\item Event Recognition Gaps:
In particular, many accidents involving pedestrians and cyclists are not recognised as significant events, with no airbag deployed, meaning no event data is stored.
  Even when EDRs activate for crashes or near-crash events, they may fail to capture events that could be considered criminal but are classified as insignificant from a regulatory perspective.

\item Authorisation and Accessibility Issues:
Ownership of EDR data is legally complex and varies by jurisdiction. For example, in the USA, ownership of the EDR data is a matter of State law, with some states considering the manufacturer to be the owner of the data, rather than the vehicle owner. While courts can obtain EDR data through court orders \cite{national2015url}, NHTSA considers the vehicle's owner to be the owner of the data collected by EDR.  
However, the accessibility of this data introduces further issues. In some cases, companies such as Tesla and BMW have recently allowed the vehicle owner to access all EDR data using some unique hardware and software. Until 2017, they were the only parties permitted to access the data \cite{tesla}. Furthermore, the original equipment manufacturer (OEM) may also access the EDR remotely in specific crash scenarios, raising concerns about who truly controls the data. \\
These legal and accessibility challenges represent obstacles to access control. Although stakeholders, such as manufacturers and vehicle owners, may require access to the data, unrestricted access presents liability risks. The need for controlled access becomes apparent, allowing stakeholders to access the data under well-defined conditions and controls could be significantly valuable. In addition, DSSAD systems require consumers' approval for recording and/or accessing their data \cite{Oica}, highlighting the need for a balanced approach to access control. Therefore, legislators, regulators, and manufacturers play a pivotal role in determining what data should be recorded, how the AI could select the valuable data, the expansion of record device capacity, and, most crucially, who should be authorised to access it.

\item Insufficient Data for Forensic Analysis:

 Current tools cannot collect all five data classes typically required for forensic investigations: firmware, communication data, user data, safety-related data, and security-related data \cite{gwehenbergerneeds}.

  A survey in which 173 international experts in accident analysis participated depicted that a considerable number of traffic accidents involving ADAS cannot be reconstructed \cite{paula2020challenges}. Furthermore, the current EDR has to be refined to provide sufficient data for liability purposes \cite{7823109}  as the current data available is insufficient for thorough forensic analysis.

\item Privacy Concerns:
 Continuous recording and storage of video, location, speed, and/or surroundings of a vehicle appear to contradict the regulations addressing privacy protection. Therefore, laws related to this matter need to be enacted before it is adopted.


\item Need for Standardisation:
Although manufacturers continue to develop new generations of EDR and DSSAD, international standards are urgently needed to ensure robust data validation and enhance their reliability. Global collaboration is needed to establish regulatory frameworks, standardise data elements, ensure interoperability, and define access authorisation.

\end{enumerate}
\subsubsection{Improving EDR and DSSAD for AV Systems}

This section explores current efforts, recommendations and solutions for improving EDR and DSSAD for AV Systems. 
\paragraph{ Recommendations/Current Efforts}
To address the limitations surrounding EDRs and DSSAD, various international efforts aim to improve current practices:

\begin{enumerate}

    \item Time Window for Recording:
    The Aggregated Homologation proposal for Event data recorder for Automated Driving (AHEAD),  which is a working group focused on developing standardized data models for recording information from a vehicle's Event Data Recorder (EDR) specifically designed for investigating accidents involving automated driving vehicles AHEAD, recommends data recording from 30 seconds before to 10 seconds after the collision. This time proves sufficient for the individual EDR-relevant claims of the present claims collective. 
    \item Inclusion of Vulnerable Road Users:
    Given the frequent involvement of pedestrians and cyclists in accidents, the Netherlands plans to immediately extend the scope of the EDR to include them.
    \item Incremental Adoption of Standards:
    The UK government proposes a two-step method. First, introducing the USA's standards for EDRs in a relatively short timeframe; then, the second step applies more extensive requirements if adopting a single-step approach is not feasible. \cite{UNECE}.
    \item Various studies in the literature highlight critical requirements for local data storage in AV, such as those proposed by Kim et al. in their works \cite{kim2022cybersecurity, kim2022data} and by Ten Holter et al. in \cite{ten2024s}. This research and similar efforts will ultimately enhance safety and accountability in AV.
    
\end{enumerate}

\paragraph{Technical Solutions:} \label{Solutions}
This section presents various solutions for accident investigations and data integrity, as summarised in Table \ref{accident_solutions} that compiles solutions from multiple studies.

Various safety analysis models, including CAST \cite{sharma2019vehicular} and FRAM \cite{patriarca2020framing}, are used for accident investigation. CAST employs system theory to identify causes of failure and propose preventive measures. FRAM assesses complex interactions in socio-technical systems. These models lack a unified framework covering all causal factors, and manual analysis remains costly and inefficient.

According to \cite{paula2020challenges}, the solution to the limited information gathered independently from the manufacturer is a Forensic Event Data Recorder (FEDR). FEDR is an EDR that meets all the requirements from the investigator’s perspective.

Data integrity preservation for investigation purposes in AV has been a goal of several studies that have presented frameworks based on various forms of technology. Hoque and Hasan have proposed a forensic investigation framework for AVs called AVGuard tool \cite{hoque2021avguard} that is designed for integration with the AD system. The framework assumes that the AV has local storage to store the log provenance while also communicating with a remote cloud server to publish the newly created log provenance. A robot operating system (ROS) node collects all the logs from different AD modules. Oham et al. have proposed a distributed digital forensics framework \cite{oham2018blockchain}, which is based on the evidence reported by nearby witness vehicles if a vehicle is involved in an accident. Digital signatures, along with a corresponding certificate, are used to protect data integrity. Data exchanges between entities in the framework are stored in a blockchain and used for later decision-making.  
Further, T-Box \cite{lee2019t} is a trusted real-time data recording system, which consists of an automotive data recording system with a network monitor, generator, and recorder. It assumes that the gateway could be used as a network monitor. The generator reconstructs data provided by the monitor and delivers it to the recorder, which stores data. The recorder stores an individual data entry into a block, and these data blocks can be stored locally or externally or transmitted to a remote server. 
Buquerin et al. \cite{buquerin2021generalized} have provided a general concept for automotive forensics. Using Ethernet, their implementation uses the onboard diagnostics interface, the diagnostics over internet protocol, as well as the unified diagnostic services for communication. Liu et al. aim to store EDR data safely and away from manipulation. In their scheme \cite{liu2022fast}, data is not only sent to the manufacturer's server as usual, but the vehicle also uploads the EDR data to a cloud server and sends the evidence of storage to the nearby vehicle through a vehicular ad hoc network.\newline
To conclude, while various solutions have been proposed to enhance accident investigations and ensure data integrity in AV systems, significant challenges remain. Issues such as data ownership, and privacy concerns still need to be addressed to develop a comprehensive and reliable approach for forensic investigations in AV environments. Future research should focus on integrating these solutions into a federated framework that balances accessibility, security, authorisation, integrity and regulatory compliance.

\begin{table}[h!] 
\centering
\caption{Solutions for Accident Investigations and Data Integrity}
\label{accident_solutions}
\renewcommand{\arraystretch}{1.1} 
\setlength{\tabcolsep}{2 pt} 
\scriptsize 
\begin{tabular}{|p{2cm}|p{2cm}|p{4cm}|p{3cm}|p{3cm}|}
\hline
\textbf{Category} & \textbf{Solution} & \textbf{Description} & \textbf{Strengths} & \textbf{Limitations} \\ \hline\hline

\textbf{Safety Analysis Models} 
    & CAST \cite{sharma2019vehicular}.
    & Uses system theory to identify failure causes and propose preventive measures. 
    & Identifies system failures systematically. 
    & No unified framework; costly and inefficient. \\ \hline
    & FRAM \cite{patriarca2020framing}.
    & Analyses complex interactions in socio-technical systems. 
    & Handles complex interactions. 
    & No unified framework; costly and inefficient. \\ \hline

\textbf{Forensic Investigation Frameworks} 
    & AVGuard Tool \cite{hoque2021avguard}.
    & Integrates with ADS, collects logs via ROS and publishes to a cloud. 
    & Supports modular log collection. 
    & Assumes reliable local storage. \\ \hline
    & Distributed Digital Forensics Framework\cite{oham2018blockchain}.
    & Relies on nearby AVs and Blockchain to ensure data integrity. 
    & Uses Blockchain for tamper-proof evidence. 
    & Requires witness AVs; may add network overhead. \\ \hline

& Automated Vehicle Data Pipeline. \cite{beck2023automated}.
    & A pipeline consists of collecting raw sensor data and processing to reconstruct crash scenarios. 
    & High-fidelity crash reconstruction. 
    & Privacy and Security concerns. \\ \hline

\textbf{Trusted Data Recording} 
    & T-Box \cite{lee2019t}.
    & Real-time data recording system with network monitoring and storage options. 
    & Reliable real-time operation. 
    & Lacks privacy considerations. \\ \hline
    & Generalized Automotive Forensics \cite{buquerin2021generalized}.
    & Uses diagnostics and Ethernet-based communication. 
    & Efficient diagnostic communication. 
    & Limited implementation details. \\ \hline

\textbf{Forensic Data Integrity} 
    & Safe EDR Storage  \cite{liu2022fast}.
    & Uploads EDR data to the cloud and shares evidence with nearby vehicles. 
    & Prevents data manipulation. 
    & Relies on vehicular ad hoc networks. \\ \hline

    \textbf{Forensic Data Integrity with Controlled Access} 
    & AVChain \cite{singh2023trusted}.
    & Blockchain and IPFS for secure, verifiable crash data sharing among stakeholders. 
    & Ensures data integrity and controlled access. 
    & The architecture's complexity and limited real-time capabilities\\ \hline

    & Forensic Event Data Recorder (FEDR)  \cite{paula2020challenges}.
    & Meets investigator requirements by gathering independent data. 
    & Enhances forensic investigation. 
    & Requires widespread implementation. \\ \hline

\end{tabular}
\end{table}


\subsection{Local Data Processing}
	The collected data feeds the onboard diagnostics, which are part of the ADS function, i.e. driving decision-making,  analysis of crashes, or technical failures.
 In addition to managing data collection from sensors and the fusion of sensor inputs, these embedded systems handle high-precision functions such as localisation, storage and updating of maps to finally perform complex tasks such as real-time control and machine learning. The data flow among these embedded systems faces many challenges, including integrating data across disparate AV systems, efficient communication \cite{elhadeedy2024autonomous}, safety, and cybersecurity challenges \cite{sonko2024comprehensive}. These challenges require advanced solutions that optimise data processing efficiency.

\subsection{Edge Processing}
There are three cloud processing approaches: centralised location, edge-based processing,  flexibility, and the Hybrid approach solution, which is the most preferred. Even though clouds are crucial for the success of AVs, there are several challenges facing the vehicular cloud community that are not faced by traditional cloud computing, such as high data transfer demands, latency concerns, and security risks. These factors drive the need for edge-based processing to offload computational tasks and reduce latency.

\paragraph{Key Challenges in Local and Edge Data Processing}
Local and edge data processing plays a crucial role in ensuring low-latency real-time decision-making, but it also presents various key challenges that must be addressed to optimise data management, performance and reliability.

\begin{enumerate}
\item Latency-Sensitive Data Processing: While the low-rate data are regularly transmitted to the manufacturer's cloud, where all processing occurs at a centralised location, the high-latency-sensitive data, however, needs high-speed data processing within a 
 distributed architecture.  
\item Dynamic Resource Allocation: Unlike traditional cloud computing, vehicular clouds face challenges due to the dynamic nature of resources.
\item Lack of Central Authority: Managing security, privacy, authorisation, and authentication is more complex without a centralised authority \cite{olariu2019survey}, \cite{kang2019autonomous}.
\item Network Bandwidth and Scalability: Large amounts of data require significant network bandwidth and scalable infrastructure for local processing\cite{ghansiyal2021information}.
\end{enumerate}

\paragraph{Solutions for Local and Edge Data Processing}
The edge-based cloud data centre performs many tasks at the edge instead of the cloud, leading to faster access than cloud computing. Edge and fog computing play a critical role here because the data requires high network bandwidth, which provides data processing and local storage capabilities. In addition, combining centralised cloud computing with edge-based solutions balances latency-sensitive and non-sensitive tasks.\\

An achievable paradigm for addressing issues with dynamic resource allocation and latency-sensitive data processing is mobile edge computing (MEC). MEC reduces reliance on centralised cloud infrastructures by bringing computational power closer to the data source, allowing quicker and more effective processing. This method works particularly effectively for ADS that must handle data securely and make decisions in real time. As summarised in Table~\ref{table2}, various studies propose solutions that optimise computation time, energy consumption, and resource usage while maximising privacy in edge environments.

\begin{table}[h!] 
\caption{Mobile Edge Computing (MEC) Solution Studies – Summary of Objectives}
\label{table2}
\centering
\begin{tabular}{|c c c c c|} 
\hline
Scheme & Minimising & Minimising & Optimising & Maximising \\ [0.5ex] 
& computation time & energy consumption & resource & privacy \\ [0.5ex]
\hline\hline
\cite{tran2018joint} & \checkmark & \checkmark & \checkmark & \xmark \\ 
\hline
\cite{liu2019deep} & \checkmark & \xmark & \checkmark & \xmark\\
\hline
\cite{8745530} & \checkmark & \xmark & \checkmark & \xmark \\
\hline
\cite{ren2019collaborative} & \checkmark & \xmark & \xmark & \xmark \\
\hline
\cite{sun2020reducing} & \checkmark & \xmark & \xmark & \xmark \\
\hline

\cite{bi2022edge} & \xmark & \xmark & \xmark & \checkmark \\
\hline

\cite{asim2023intelligent} & \xmark & \checkmark & \xmark & \xmark \\

\hline
\cite{7931658} & \checkmark & \xmark & \xmark & \checkmark\\
\hline
\cite{liang2021distributed} & \checkmark & \xmark & \checkmark & \xmark \\[1ex] \hline \hline
\end{tabular}
\end{table}


\subsection{Backend Cloud Computing}

Most companies acknowledge and state that they keep a lot of data regarding the vehicle owner, the vehicle itself, and its in-vehicle hardware and software products \cite{tesla}. 
The period of retention of this data varies from one company to another. They also differ in how they obtain this data; some transfer it physically to backend servers, while others do so remotely via networks. In addition, they differ in the data format; companies such as Tesla use raw sensor data, whereas some other companies ask Original Design Manufacturers (ODM) to provide processed data.\\
Captured and stored data requires substantial storage infrastructure, for instance, cloud or on-premises servers. The majority of automakers are utilising cloud-based capabilities via connected-car services. For example, in 2019, Ford publicised its connected-vehicle collaboration with AWS. In addition, Toyota introduced its engineering ecosystem in 2020 to develop and deploy the next generation of cloud-connected vehicles alongside similar initiatives outlined in \cite{SAE}. \newline
 Indeed, in a promising attempt to address the data management challenges in AVs, Cloud technology is a scalable solution in the automotive industry. Therefore, data is classified as onboard data, sent to the cloud or stored in servers and hard drives as long-term storage. Local function data and V2X data that has a 4ms response time requirement and has to go off-vehicle will not be sent to the cloud \cite{experts}. In contrast, model training data, for example, includes cases where a new object has been detected, and the data about this anomaly will be used to formulate patterns and reports will be sent to the cloud for future algorithm improvements \cite{carcloud}, \cite{wang2019multi}.

\subsubsection{Key Challenges and Current Solutions}

 The following key challenges in backend cloud computing for AVs appear insufficiently addressed in the literature.

\begin{figure}[H]
  \centering
  \includegraphics[scale=1]{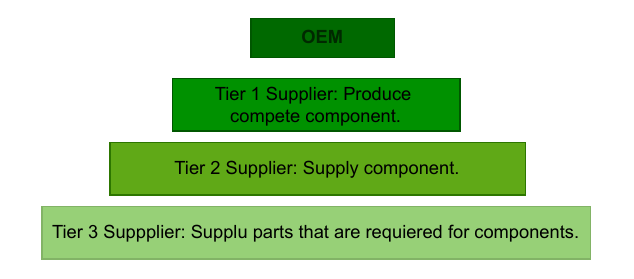}
  \caption{Hierarchy of supplier tiers in automotive manufacturing.}
  \label{The Levels of Tiers of Any Manufacturer}
\end{figure}

\paragraph{1. The Need for Data Sharing}
Even though some vehicle manufacturers agree to share in-vehicle data with the other service providers by accessing data directly through the vehicle manufacturer’s server or via “neutral” servers that would gather the data, the service providers ask for direct real-time access to in-vehicle-produced data and functions through an in-vehicle interoperable, standardised, secure, and open-access platform \cite{geradin2020access}. In addition to legal authorities, Tesla states that the data could be shared with their service providers, business partners, and affiliates; in addition to any third parties, the owner has also been authorised \cite{tesla}. Some stakeholders suggest that car manufacturers should allow tier 1 suppliers to access the data directly \cite{mccarthy2017access} to maintain and improve their products. Figure~\ref{The Levels of Tiers of Any Manufacturer} illustrates the tier levels within a manufacturer.

Various situations are defined in the literature where stakeholders need to access accurate AV data, such as accident and failure data.
In traditional vehicles, human error is the critical cause of 94\% of vehicle accidents~\cite{singh2015critical}. However, in the case of an accident with an AV, there are clear differences in liability. In particular, the interaction of many factors in AVs and their ability to make some or all driving decisions disrupts the adjudication process. The liability subjects of AV accidents may include vehicles, vehicle assistant drivers, manufacturers, vehicle owners, and insurance companies~\cite{yuan2021key}. However, the need for this data from stakeholders encompasses a much broader scope than these parties.
 In the literature, authors and organisations differed in their identification of those stakeholders and the techniques they use to identify them. NHTSA states that motor vehicle accidents may be investigated through various entities, and they listed them \cite{correia2001utilizing}. Another example, the authors in ~\cite{omeiza2021explanations} fragmented stakeholders into three general classes: A) containing all types of end-users and society; B) containing all technical groups; and C) for all regulatory parties, including insurers. However, the most frequent entities in literature are as follows:
\begin{enumerate}
    \item Government and the legal authority: The accident data collected from the involved AV are necessary as facts and evidence in criminal courts to determine liability. Another possible scenario for government use is to take advantage of data to reduce the possibility of accidents in the future and thus reduce losses and assess roadside safety.
    \item Original Equipment Manufacturer (OEM): The Manufacturers may analyse the collected data and use it to monitor system performance and improve their products. Another possible scenario is to reconstruct the accident to assess the causes in order to reduce the possibility of accidents in the future.
    \item Suppliers: The potential scenario is to analyse and improve the products or services that are involved in a vehicle manufactured by the OEM.
    \item The owner: In one possible scenario, the owner might use the data as evidence to absolve him/her of any criminal liability. In addition, the provided data could be used for insurance amounts and services.
    \item Insurance: The stored data presents accurate evidence to resolve insurance disputes and a fair solution that ensures that no party is tampered with. 
    \item Testing and certification bodies: The testing organisation needs the data to reconstruct the accident in order to analyse and update the technical and legal requirements.
    \item Road authorities: Data can support assessing and improving the infrastructure and roadside safety.
    \item Researchers: To analyse and assess the AV crashes to assist in the development of vehicles, infrastructure, and the whole environment’s components.

In addition, based on this study's findings, a further entity has been identified:

\item Other Manufacturers: This category includes companies involved in manufacturing the vehicles. These manufacturers can leverage the collected data to improve their products, assess compatibility with AV technologies, and enhance training for AVs, especially when they encounter new cities or environments they have never experienced before. This strengthens the AV’s adaptability and operational efficiency, contributing to the development of safer and more effective ADS.

\end{enumerate}
Due to the significance of preserving the data generated and received by the system, it is clear that sharing this data with all relevant parties is essential. 

\paragraph {Current Solutions}

 One approach that facilitates collaborative data sharing and mitigates privacy concerns is using Federated Learning (FL), which is a distributed collaborative AI approach that allows multiple devices to coordinate data training with a central server without sharing actual datasets \cite{nguyen2021federated}. With a trusted server (aggregator), parties can learn a shared machine learning model locally and separately, as well as share only the resulting insights from each analysis. This technology is an active area adopted recently in various applications and research such as in financial applications \cite{byrd2020differentially}, mobile applications \cite{bonawitz2017practical}, biomedical research \cite{froelicher2021truly}, which relies on Multiparty Homomorphic Encryption (MHE) to perform privacy-preserving FL by using the advantages of both interactive protocols and homomorphic encryption (HE). Another example is to encrypt only the critical parts of model parameters to reduce local computation and communication costs. Sotthiwat et al. ~\cite{sotthiwat2021partially} proposed a partially encrypted Multi-Party Computation (MPC) solution that only encrypts the first layer of local models with MPC strategy.  

In the AV domain, blockchain-based FL systems have been recently introduced to enhance security, transparency, and reliability in data sharing. For example, a distributed on-vehicle machine learning model \cite{pokhrel2020federated} has been proposed to improve vehicular networks. Despite the increased computational overhead in that work, data could be trained, and models could be exchanged in a distributed manner. Similarly, a framework has been presented to enable vehicles to encrypt portions of their data using HE before uploading it to a cloud server\cite{nguyen2024towards}. Zeng et al \cite{zeng2022federated} also designed an automated controller to avoid performance deficiencies of traditional learning-based controllers that are trained by each connected vehicle’s local data. In their design, the learning models used by the controllers are collaboratively trained among a group of CAVs.\newline
While these approaches and similar ones demonstrate potential solutions, they require further refinement to enable AV manufacturers to collaboratively benefit from FL algorithms by sharing knowledge across their respective clouds without exchanging any raw data. Moreover, they could be effectively integrated to facilitate collective training that would otherwise be unattainable individually.

On the other hand, while these algorithms preserve privacy, this advantage comes at the expense of the model's accuracy due to encryption and the limited operation set they support. In other words, approaches such as  FL, HE, and MPC offer significant advantages in preserving data privacy and security, and play a crucial role in facilitating secure data analysis. However, in certain use cases, sharing partial or complete datasets is a desirable goal to enhance model accuracy and performance, allowing for more comprehensive insights and better decision-making while protecting sensitive information.

ADS in AV as an emerging and future-oriented field requires broader and more effective collaboration and data sharing among the stakeholders mentioned in the previous section. Thus, there is an urgent need for systems that integrate highly secure data sharing for AV. The literature presents several attempts to propose secure access control mechanisms in various data fields, such as ~\cite{liu2021secure}, \cite{zhao2022attribute}, \cite{bianchi2019intelligent}. Even though there are a few notable systems that have been developed and tailored to AV data \cite{singh2023trusted, alhabib2024hyperledger}, similar works in this field are rare and nearly nonexistent.

\paragraph{2. Data Validation}
The challenges of data sharing are not limited to direct access; rather, there is also a need to validate data. However, due to the large volume of heterogeneous data, it is quite difficult to even pre-process it effectively. Therefore, data validation becomes a critical step in this stage of the data lifecycle to ensure consistency, accuracy, and reliability.

\paragraph{Current Solutions}

A blockchain-based platform with MPC is proposed for the AV data validation process \cite{khan2024believe}. BELIEVE, which stands for Blockchain-Enabled Location Identification and Efficient Validation with Encryption approach, integrates real-time data sharing for immediate decision-making with backend storage on a distributed ledger to ensure that validated data is securely recorded and accessible for future reference. AV systems rely heavily on big data analytics, thus, data quality improvement strategies are critical to address challenges relevant to a big data environment, such as the management, storage, cleaning, integration, and reducing inconsistencies and optimisation. Data quality improvement studies have been proposed recently, such as \cite{ pillai2024techniques, zuo2023research, ahmad2023analysis}. These studies and further programmes are needed as essential parts of validating data quality before and after it is stored.

%% file: Section4_concerens.tex
\section{Cross-Cutting Concerns}\label{Concerns}

The ADS operate in a complex and dynamic environment, generating vast amounts of data that must be processed, stored, and shared seamlessly. This section explores certain overarching concerns that impact the system as a whole.  Addressing these concerns is critical to ensure reliability, trustworthiness, and ethical operation.

\subsection{The Complexity of Fault Detection}
An ADS generates a vast amount of data, making the extraction of crucial information from the logs of different AD modules a real challenge.  In the case of a defect, finding the set of influencing factors causing the failure is a complex mission due to two reasons: first, the AD functions may not be sufficient for all unexpected conditions in the dynamic environment with unlimited contexts; second, deviation from the intended functionality due to the inductive nature of a system that combines machine learning components. Failure in sensor fusion, sensor readings (e.g. misdetection), external environment context (e.g.: weather), or control issues (for example: braking is not initiated in time), can result in unintended outcomes, sometimes due to combinations of these factors \cite{zhang2021finding, neurohr2021criticality}. Another proposed taxonomy by Zhao et al. \cite{zhao2024potential} for potential sources of sensor data anomalies in AV is categorised into four main categories: faults in components, adaptability failures, cyber-attacks, and design deficiencies.
Within the enormous amount of produced data, locating meaningful data that pinpoints the crash cause or causes is a significant challenge. Identifying the precise moment of failure, or the faulty device or subsystem, adds further complexity. In discussing the complexity of fault detection, various external factors contribute to scenarios that may lead to dangerous situations. Identifying these causes of harm complicates fault detection because it requires a broader understanding of the traffic context, not just the internal functioning of the vehicle. 

Figure~\ref{The Complexity of Finding The Source of Harm}  illustrates the complexity of sources of harm in the ADS based on ISO 26262, ISO/PAS 21448, and SAE J3016 standards. It categorises harm arising from malfunctioning behaviours, functional insufficiencies, and deviations from the Operational Design Domain (ODD), the specific conditions where the ADS is designed to operate safely. In addition, it highlights that even with proper design intent, failures can occur, and the operational context influences these sources of harm. Furthermore, since crashes are rare events, testing and analysing similar scenarios to define the exact harm that may be blamed on one party or another in this nested system is particularly challenging. 

\begin{figure}[H]
  \centering
  \includegraphics[scale=0.62]{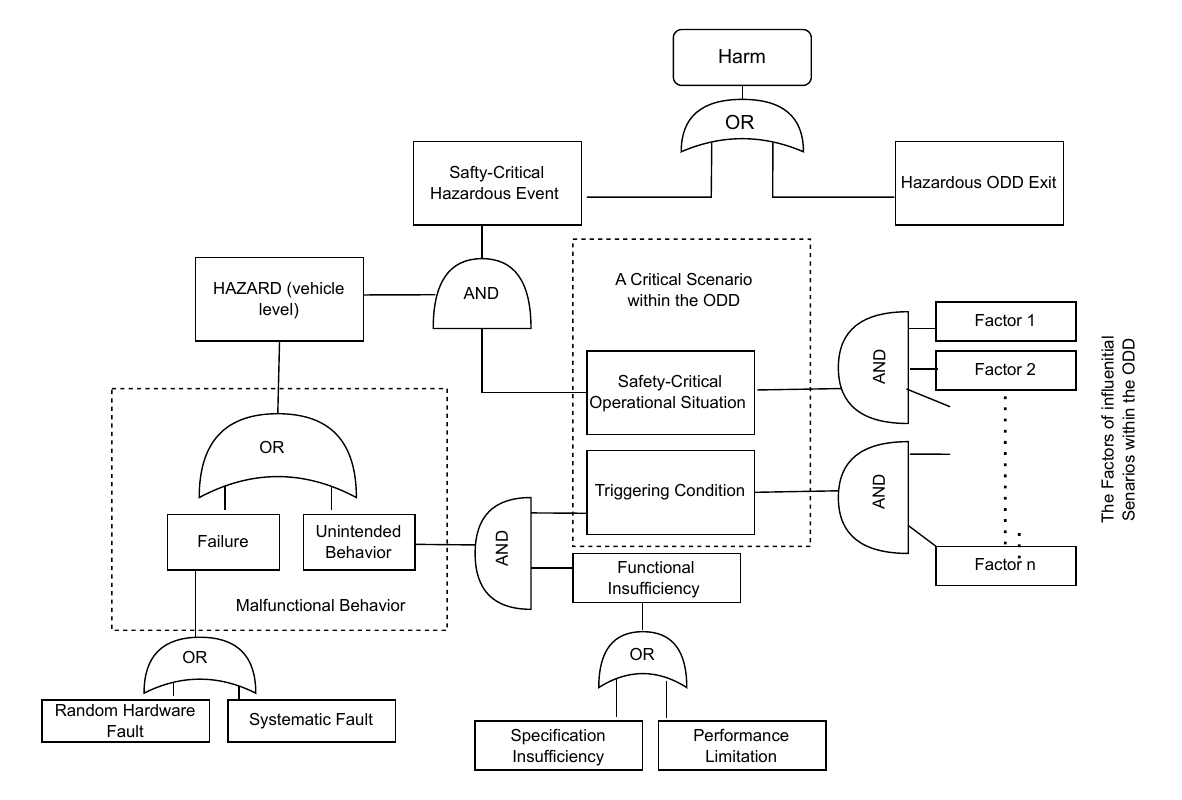}
  \caption{The Complexity of Finding The Source of Harm\cite{zhang2021finding}.}
  \label{The Complexity of Finding the Source of Harm}
\end{figure}
For real-time safety assessment,  \cite{kamel2024real} has proposed the Bayesian Hierarchical Spatial Random Parameter Extreme Value Model (BHSRP) for real-time safety assessment. This model could address the difficulty of extracting meaningful information from the massive and complex datasets generated by ADS.

In this context, a conceptual fault-handling system design for driverless trucks was proposed in \cite{rylander2024conceptual}, which highlights the need for advanced fault detection mechanisms in driverless vehicles. The findings in the paper emphasise the difficulty in determining the exact cause of a failure. Similarly, Koopman and  Wagner \cite{koopman2016challenges} discuss the main challenge in testing AV, framing their analysis within the V-model developed under ISO 26262. This framework links various types of testing to ensure that safety-critical systems meet required standards. However,  ISO 26262:2011, which sets standards for safety-critical systems, requires adaptation to accommodate the complexities of future autonomous driving technologies \cite{griessnig2017development}. In the same context, A criticality analysis framework \cite{neurohr2021criticality} aimed to identify and analyse critical traffic phenomena for the verification and validation of automated vehicles.  The proposed mechanism involves a combination of both expert-based and data-driven approaches to identify relevant criticality phenomena and explain the underlying causes. These studies and similar ones lack a comprehensive approach to identifying influencing factors in such a complex and dynamic traffic environment. They also, as mentioned earlier, face limitations in model constraints and data availability when assessing critical situations.

Maintaining the high-quality training data sets is crucial for AI systems that support decision-making in AV \cite{Europe}. Predictive models that leverage machine learning approaches can further enhance data validation by classifying useful and non-useful data, thereby improving real-time analytics. For example, the machine learning-based approach for predictive analytics, as in \cite{goriparthi2024ai}, offers possible solutions for identifying system failures, optimising data validation and addressing sensor misdetection and control failures in such dynamic environments. 
Further research is needed to address the challenges of identifying the source of harm and fault in AV due to the complexity of their systems and operating environments.

\subsection{Data Privacy}
The open wireless access in VANET seriously impacts not only the privacy of users, but also the pedestrians whose photos and locations can be captured by vehicles. AVs generate a vast amount of data that is collected, stored, transferred, and shared, posing severe risks to user privacy.

Privacy concerns span all phases of data flow in AVs and require thorough investigation. Table \ref{tab:privacy-concerns} summarises key studies addressing these challenges, detailing their aims, mechanisms, results, and limitations in providing privacy solutions for AV.


\begin{table*}[!h]
\centering
\caption{Summary of Privacy Concerns and Solutions for Autonomous Vehicles in the Literature}
\resizebox{\textwidth}{!}{%
\begin{tabular}{|p{2cm}|p{3.3cm}|p{3.3cm}|p{3.3cm}|p{3.3cm}|}
\hline
\textbf{Paper} & \textbf{Aim} & \textbf{Mechanism} & \textbf{Results} & \textbf{Limitations} \\
\hline

\cite{9945637}
 &To preserve privacy in AV using homomorphic encryption during the storage and processing stages. & A Pixel-level encryption for secure searching over encrypted images with probabilistic trapdoors.& Reduces storage and increases efficiency. &Lacks detailed discussion on real-world implementation and performance impacts on various cloud environments. \\
 \hline

\cite{zhang2021efficient} & To develop a privacy-preserving authentication scheme for V2G networks. & The scheme uses randomly selected pseudonyms for AVs and establishes a secure session key through an authentication key agreement protocol to ensure confidentiality. & The proposed scheme achieves lower communication overhead (800 bits) compared to existing schemes. & The paper does not provide real-world testing of the proposed scheme. \\
\hline

\cite{parekh2023gefl} & Enhance privacy in federated learning for AVs
& Gradient encryption in federated learning to preserve user privacy without extra computational cost.  & Improved accuracy (2\% higher) and reduced data transfer compared to conventional FL. &  Additional computational infrastructure needed for blockchain integration. \\
\hline

\cite{gheisari2024cappad} & Propose a context-aware privacy-preserving method for AVs.  & Uses SDN and differential privacy/data aggregation depending on data sensitivity. & Shows higher performance in privacy preservation, cost, and latency compared to existing methods. & May require further evaluation with third-party providers; potential limitations in highly dynamic environments; computational complexity and overhead considerations. \\
\hline

\cite{10.1145/3583740.3628436} & Develop a framework that allows users to choose which parts of their data they want to keep private before sharing it with other vehicles.
& PRECISE framework utilises secure segmentation to identify sensitive objects, inpainting techniques to remove them, and edge computing to process data using secure deep learning models enhanced by additive secret sharing.
& The framework achieved secure segmentation in 3.47 seconds and inpainting in 0.99 seconds.
& Processing times may impact real-time performance, and the security of edge servers poses a risk of data exposure.

\\
\hline

\cite{zhao2024privacy} & To improve traffic efficiency and fuel economy while protecting the privacy of vehicle data using cloud-based collaboration.
& an affine masking-based privacy strategy, which encrypts vehicle state data before sending it to the cloud and decrypts the control input using inverse affine masking to protect privacy during vehicle-cloud collaboration.
& The scheme enhances traffic efficiency and fuel economy while securely masking vehicle data through an affine masking technique
& The limitations include potential computational overhead affecting real-time performance and scalability challenges when handling a large number of vehicles simultaneously.

\\
\hline

\end{tabular}%
}
\label{tab:privacy-concerns}
\end{table*}


\subsection{Security}

Security in AV is a critical concern as network vulnerabilities can lead to severe consequences, including loss of life. Due to the complexity of AV systems, security threats target various components, and these attacks can be broadly classified into the following categories: \newline

\paragraph{In-vehicle network}

Attacks on In-vehicle networks include the following:
\begin{enumerate}

    \item \textbf{ECU engine control units:} Attackers may compromise the ECU by altering programming code, affecting vehicle performance.
    \item \textbf{CAN and SEA J1939 buses:} The CAN bus, which connects all the vehicle’s components, is a critical target for attacks. For example, malicious actors can inject viruses into the CAN bus, disrupting critical operations
    \item \textbf{Remote sensors:}  Any tampering with the sensors' data generated and transmitted can result in fatal accidents. Through existing wireless networks, external entities can make connections with sensors, implementing remote sensor control.
    \item \textbf{GPS:}  Adversaries can alter GPS data, disrupting navigation and decision-making processes.
    \item \textbf{Wireless communication:} Wireless technologies such as Bluetooth, tire pressure monitoring systems (TPMS), and keyless entry and ignition systems present additional vulnerabilities. Attackers may exploit these systems to gain unauthorised access or control over the vehicle \cite{sun2021survey}.\newline
\item \textbf{Denial-of-Service (DDoS):} Distributed Denial-of-Service (DDoS) attacks have emerged as a critical concern within the V2X and VANET environments. These attacks flood vehicle communication channels with malicious traffic, leading to service disruptions, degraded safety performance, or even system failure.

\end{enumerate}

\paragraph{Vehicle to everything network (V2X)}

\begin{figure}[H]
  \centering
  \includegraphics[scale=0.98]{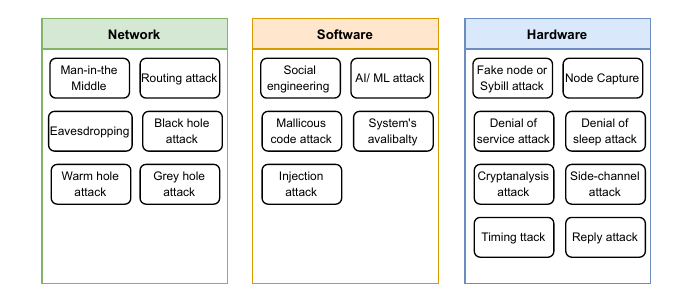}
  \caption{Attack Taxonomy \cite{gupta2022survey}.}
  \label{Attack Taxonomy}
\end{figure}

Attacks on V2X communications typically target the following systems:\begin{enumerate}

    \item \textbf{VALNET or Vehicle ad-hoc networks:} VANETs rely on dedicated short-range communications (DSRC) and are based on the IEEE 802.11p standard for wireless access in vehicular environments.
    \item \textbf{Mobile Cellular Network, Satellite Radio, and Bluetooth:} Another communication structure required for V2X are the mobile cellular network, satellite radio, and Bluetooth, which can be targeted by attackers to disrupt vehicle communication or gain unauthorised access to vehicle systems.

\end{enumerate}
Another taxonomy of attacks has been provided by Gupta et al. \cite{gupta2022survey} based on the architecture of hardware, network, and software as in Figure~\ref{Attack Taxonomy}.

Several solutions in the vehicular network have been recently proposed. For instance,  RTED-SD, which is a scheme that aims to detect real-time attacks \cite{li2021rted}. In this scheme, the authors use the Fast Quartile Deviation Check algorithm (FQDC) to recognise and locate the attack in the Internet of Vehicles. Similarly, a threat prevention framework has been proposed in \cite{anwar2024dynamic} for Vehicle-to-Vehicle communication in AV Networks, integrating dynamic risk assessment using the Probability-Impact-Exposure-Recovery (PIER) metrics, security decay assessment via ruin theory, and a risk-aware message forwarding algorithm based on game theory. This approach aims to enhance security and privacy by proactively addressing vulnerabilities in V2V communication.
Furthermore, to detect malware attacks in AV, Aurangzeb et al. \cite {aurangzeb2024cybersecurity} proposed a hybrid approach that combines static and dynamic analysis for real-time detection. The mechanism demonstrated malware detection that also enhanced safety and reduced communication latency. Additionally, a method proposed by Cretu et al. \cite {cretu2022querysnout} is a method that utilises evolutionary search and machine learning to identify vulnerabilities in query-based systems (QBS). It showed higher performance in finding vulnerabilities compared to existing attacks.

Various works have explored and enhanced Intrusion Detection Systems (IDS) in AV, aiming to address different challenges and approaches. Anthony et al. \cite{anthony2024intrusion} propose a method called NTB-MTH-IDS, a Non-Tree-Based Multi-Threshold Hybrid Intrusion Detection System, which is an intrusion detection system that leverages non-tree-based machine learning techniques. Similarly, Anbalagan et al. \cite{anbalagan2023iids} introduce IIDS, which is based on a deep Convolutional Neural Network (CNN) system that transforms vehicular network traffic data into images for attack detection. Additionally,  Aloraini et al. \cite{aloraini2024adversarial} investigate adversarial attacks on IDSs in in-vehicle networks.
For detecting DDoS attacks in VANET, recent studies have investigated the machine learning-based techniques, for example, the approach presented by Setia et al. \cite{ setia2024securing}.

Another security approach for AV that has emerged as a promising solution is quantum encryption, which aims to face the potential threats posed by quantum computing.  Despite the complexity of quantum-based security solutions, several works have explored their application in AV. For instance,  including blockchain-based authentication, quantum key distribution, and secure federated learning \cite{10613753,10287202,10061684}.

Various methods recently have proposed the integration of BC technology in IoV systems, aiming to enhance security. For example, \cite{dwivedi2020blockchain} uses a consensus mechanism to provide a secure event-sharing protocol for smart cities using IoV and RSU. Similarly, \cite{yi2022secure} aims to protect IoV against quantum attacks. In addition, a blockchain-based mechanism  \cite{liu2020b4sdc} is designed to provide security-related data collection and incentivise mobile nodes.

\begin{table*}[!h]
\centering
\caption{Summary of Security Solutions for Autonomous Vehicles in the Literature}
\resizebox{\textwidth}{!}{%
\begin{tabular}{|p{3.3cm}|p{3.3cm}|p{3.3cm}|p{3.3cm}|p{3.3cm}|}
\hline
\textbf{Security Solution} & \textbf{Application Area} & \textbf{Mechanism} & \textbf{Results / Contribution} & \textbf{Known Limitations} \\
\hline

RTED-SD (Real-Time Event Detection-Stop Detection) & Road event detection, Driver behaviour analysis & Hypergraph-based technique for analysing multi-vehicle interactions to detect risky events in real time & Enables fast detection of hazardous driving scenarios & High computational cost due to complex multi-vehicle data \\
\hline

Hybrid Malware Detection & Malware detection in V2X communications & Combining static and dynamic malware analysis in real-time systems & Enhancing malware detection while maintaining safety and reducing communication latency & Processing overhead and real-time complexity \\
\hline

Intrusion Detection Systems (IDS) & Detection of DDoS, adversarial, and spoofing attacks in vehicular networks & ML-based IDS (e.g., NTB-MTH-IDS), CNN-based traffic analysis, GAN-based adversarial attack detection & Improves detection of varied cyber threats using adaptive learning & High dependence on training data; costly model updates; computational intensity \\
\hline

Quantum Encryption & Secure communication and confidentiality & Uses quantum key distribution and federated learning for AV communications & Offers future-proof security resistant to quantum threats & High implementation complexity and cost of quantum hardware \\
\hline

Blockchain-based Authentication & Data integrity, privacy, and access control in V2X & Uses smart contracts and distributed ledgers for secure authorisation and tamper-proof logging & Enhances trust and transparency in AV systems & Scalability issues and computational overhead in high-speed networks \\
\hline

Blockchain-based Incentive Mechanism & Secure data sharing and collaboration in IoV environments & Incentivises honest behaviour in AV networks through token-based mechanisms linked to data integrity & Encourages cooperation while ensuring verifiability of shared data & Assumes predefined threat models; integration complexity \\
\hline

\end{tabular}%
}
\label{security_solutions_av}
\end{table*}

Despite the advancements of these solutions, they still face notable limitations. In addition to their computational overhead and integration complexity, many approaches rely on predefined threat models, limiting their ability to detect unknown attacks. An overview of the key security solutions proposed for autonomous vehicles, including their applications, approaches, and limitations, is presented in Table \ref{security_solutions_av}. A thorough understanding of both vulnerabilities and potential solutions is essential to advance the security and privacy of AV in a connected landscape.


\subsection{Regulation and Standards}

Most of the current regulations that relate to human drivers must be changed to properly address AV, including testing and deployment, safety standards, data exchange standards, security standards, liabilities, insurance, and personal information privacy standards. For example, much of the current privacy legislation is inappropriate for AVs, such as the U.S federal Drivers’ Privacy Protection Act, and Electronic Communications Privacy Act \cite{UNECE}.\\
Indeed, some countries have started acting on new regulations; for instance, the Scottish Law Commission’s regulation for AV and AV legislation and policies in the USA, the Netherlands, the UK, and Sweden \cite{omeiza2021explanations},\cite{faisal2019understanding}. In addition, EU countries, industry, and the Commission are collaborating to achieve the EU's ambitious vision for connected and automated mobility across the EU. The commission will address many current issues, such as policies and legislation relating to digital technology, including cybersecurity, liability, data use, privacy, and radio spectrum/connectivity, which are of increasing relevance to the transport sector \cite{EuropeanUnion}. This supports the aforementioned 5G PPP; however, there are no international central regulatory bodies existing to regulate the implementation and deployment of AV.  In light of these regulatory challenges, various global principles can offer valuable insights into shaping the future of AV regulation, such as
the Precautionary Principle (PP), the Principle of Preventive Action, and the Best Available Techniques (BAT) Principle. 
 There is a growing need for greater academic investigation into the best legal and regulatory practices for ADS.

 Recent studies \cite{resnik2024precautionary, poysti2024precautionary, Ten} have suggested frameworks and approaches to guide the development and regulation of AVs and their related technologies. Similar and further efforts are needed to help balance the benefits and risks of AV by incorporating safety and ethical considerations into policies and standards.

\subsubsection{Cross-Border Interoperability}
Accessing diverse services and sharing data among vehicles and infrastructure is a critical component of vehicle decision-making. However, interconnectivity and interoperability remain significant challenges in our geopolitically partitioned World. Cooperative Intelligent Transportation Systems (C-ITS), a paradigm that is based on Information and Communication Technologies (ICT), enables the creation of both stand-alone in-vehicle systems and cooperative systems (V2X) \cite{lu2018c}. While C-ITS solutions provide valuable services, the deployment of their infrastructure and delivery of their services often encounter territorial and regulatory hurdles \cite{vlacic2022cross}. 
An architecture has been presented in \cite{naranjo2021cross} that addresses these challenges by focusing on the integration of cooperative intelligent transportation systems in automated driving with an emphasis on cross-border interoperability. This AUTOCITS  architecture was implemented in three European cities, showing the potential for harmonized systems across national boundaries. Achieving seamless cross-border interoperability in the ADS of AV requires significant collaboration among stakeholders, including governments, industry, and international standardization bodies. Bridging regulatory and technical gaps across regions remains essential for realising the full potential of C-ITS and AV ecosystems.

%% file: Section5_results_and_conclusion.tex
\section{ Conclusion and Future work}\label{Future}

This paper highlights the essential role of data authorisation and validation in Autonomous Vehicle ecosystems, reviewing recent advances at each stage of the data lifecycle. As AV technology evolves, traditional data frameworks must adapt, with technologies like blockchain, zero-knowledge proofs, and federated learning playing increasingly vital roles. Global regulatory collaboration and integrated systems are also essential to ensure the safety, privacy, and security of AVs and their connected infrastructure. The key findings and insights of this review are as follows:
\begin{enumerate}
    \item Data Flow, Integrity, and Accessibility:
Secure data management is fundamental for ADS. The flow of data, from acquisition and processing to sharing and storage, must be protected from unauthorised access and manipulation to ensure system reliability and safety. The integrity of this data is critical, not only for real-time decision-making but also for accident reconstruction, forensic investigation, liability, production development, research, regulatory,  and for all parties involved in this operation cycle. Modern technologies such as Blockchain can provide immutable records for data validation, ensure traceability in data exchanges, and enhance transparency in AV systems at each stage. However, in most frameworks, it does not inherently provide access control or authorisation, which must be handled by other mechanisms.
\item	Data Ownership, Ethical Considerations, and Authorisation:
The data ownership in AVs is complex, with multiple stakeholders, including manufacturers, governments, and users, each having a vested interest in the data generated by these systems. This complexity presents obstacles in sharing solutions due to ethical considerations and conflicting interests. Therefore, any data authorisation process must ensure that only authorised entities can access or modify critical vehicle data to strike a balance between innovation and privacy concerns. Multi-factor authentication (MFA) strengthens identity verification, zero-knowledge proofs (ZKPs) allow verification without exposing data, secure multi-party computation (SMPC) enables computations on encrypted data, and federated learning supports decentralised data training without direct sharing. Combining these approaches can enhance access control while preserving privacy. C
\item Regulatory Challenges and Cross-Border Interoperability:
Existing regulatory frameworks are primarily designed with human drivers in mind and still do not fully adequately address the unique challenges posed by AVs, particularly in terms of data privacy, cybersecurity, liability and data ownership. With AV technology rapidly evolving, new regulations must be enacted to address these issues effectively. There is also an urgent need for international collaboration to establish standards that maintain and ensure data security and privacy in different regions.
\item  Safety and Security Risks: AV systems face growing cybersecurity threats, requiring a multi-layered defence. Approaches such as Intrusion Detection Systems (IDS) and anomaly detection use AI-driven techniques to identify real-time cyber threats. In addition, Quantum-safe cryptography is crucial for protecting AV systems against future quantum computing attacks. Furthermore, there is an urgent need for threat prevention frameworks that incorporate risk assessment models and game theory to help proactively further mitigate potential cyber risks. By integrating these technologies, AV systems can strengthen security measures while ensuring safe and efficient driving operations.

\end{enumerate}

Future work should focus on collaborative efforts among industry stakeholders, regulators, and researchers to build scalable, compliant, and secure data authorisation and validation frameworks for AVs.

\subsection*{Acknowledgements}
The authors thank the support of the Saudi Arabian Cultural Bureau, Shaqra University, the University of York, and the EPSRC.